\documentclass[letterpaper, 10 pt, conference]{ieeeconf}  % Comment this line out if you need a4paper

\IEEEoverridecommandlockouts % This command is only needed if 
\usepackage{graphicx}
\usepackage{cuted} % 支持双栏中插入跨栏内容
\usepackage{caption}

\usepackage{booktabs}
\usepackage{multirow}
\usepackage{multicol}
\usepackage{makecell}
\usepackage{array}
\usepackage{amssymb}
\usepackage{amsmath}
\usepackage{amsfonts}
\usepackage{xcolor}
\usepackage{nicefrac}

\IEEEoverridecommandlockouts
\title{\LARGE \bf
Systematic Characterization of Minimal Deep Learning Architectures: A Unified Analysis of Convergence, Pruning, and Quantization
}

\author{Ziwei Zheng$^{1}$, Huizhi Liang$^{1}$, Vaclav Snasel$^{2}$,\\  Vito Latora$^{3}$, Panos Pardalos$^{4}$, Giuseppe Nicosia$^{5}$ and Varun Ojha$^{1}$% <-this % stops a space
\thanks{*This work was not supported by any organization}% <-this % stops a space
\thanks{$^{1}$Ziwei Zheng, Huizhi Liang, and Varun Ojha are with the School of Computing, Newcastle University, Newcastle, UK {\tt\small z.zheng2, huzihi.liang,  varun.ojha@newcastle.ac.uk}}%
\thanks{$^{2}$Vaclav Snasel is with VSB-Technical University of Ostrava, Ostrava, Czech Republic
{\tt\small vaclav.snasel@vsb.cz}}%
\thanks{$^{3}$Vito Latora is with Queen Mary University of London,  London, UK
{\tt\small v.latora@qmul.ac.uk}}%
\thanks{$^{4}$Panos Pardalos is with University of Florida, Florida, USA {\tt\small pardalos@ufl.edu}}%
\thanks{$^{5}$Giuseppe Nicosia is with University of Catania,  Catania, Italy {\tt\small giuseppe.nicosia@unict.it}}%
}

\begin{document}
%\title{On Phase Transition in Deep Neural Networks}
%\title{Convergence Profiles and Pruning Analysis of a Class of Deep Neural Networks: the Minimal Architecture for Stable Learning}

%\title{Minimal Neural Networks for Stable Learning by Convergence, Pruning and Quantization Analysis}
% Convergence and Pruning Analysis of a Class of Deep Neural Networks

% \author{
% \IEEEauthorblockN{Anonymous Author}\\
% \IEEEauthorblockA{\textit{Anonymous Affiliation} \\
% anonymous email}
% }

% \author{
% \IEEEauthorblockN{Ziwei Zheng}\\
% \IEEEauthorblockA{\textit{Newcastle University} \\
% Newcastle, UK \\
% z.zheng21@newcastle.ac.uk}
% \and
% \IEEEauthorblockN{Huizhi Liang}\\
% \IEEEauthorblockA{\textit{Newcastle University}\\
% Newcastle, UK \\
% huizhi.liang@newcastle.ac.uk}
% \and
% \IEEEauthorblockN{Vaclav Snasel}\\
% \IEEEauthorblockA{\textit{Technical University of Ostrava}\\
% Ostrava, Czechia\\
% vaclav.snasel@vsb.cz}
% \and

% \IEEEauthorblockN{Vito Latora}\\
% \IEEEauthorblockA{\textit{Queen Mary University of London}\\
% London, UK \\
% v.latora@qmul.ac.uk}
% \and
% \IEEEauthorblockN{Panos Pardalos}\\
% \IEEEauthorblockA{\textit{University of Florida}\\
% Florida, USA \\
% pardalos@ufl.edu}
% \and
% \IEEEauthorblockN{Giuseppe Nicosia}\\
% \IEEEauthorblockA{\textit{University of Catania}\\
% Catania, Italy\\
% giuseppe.nicosia@unict.it}
% \and
% \IEEEauthorblockN{Varun Ojha}\\
% \IEEEauthorblockA{\textit{Newcastle University}\\
% Newcastle, UK \\
% varun.ojha@ncl.ac.uk}
% % 
% }
%\makeatother
\maketitle
%\thispagestyle{empty}
%\pagestyle{empty}

% % 使用 strip 环境插入跨栏图片
% \vspace{-10em} % 减少标题和图片之间的空白
% \begin{strip}
%     \centering
%     \includegraphics[width=0.98\textwidth]{plot/abstract_img.png}
%     \captionof{figure}{Example 3-layer DNN architecture design and its performance phase analysis, including pruning and quantization.}
%     \label{fig:abstract-figure}
%     %\vspace{10pt}
% \end{strip}

\begin{abstract}
Deep learning networks excel at classification, yet identifying minimal architectures that reliably solve a task remains challenging. We present a computational methodology for systematically exploring and analyzing the relationships among convergence, pruning, and quantization. The workflow first performs a structured design sweep across a large set of architectures, then evaluates convergence behavior, pruning sensitivity, and quantization robustness on representative models. Focusing on well-known image classification of increasing complexity, and across Deep Neural Networks, Convolutional Neural Networks, and Vision Transformers, our initial results show that, despite architectural diversity, performance is largely invariant and learning dynamics consistently exhibit {\it three regimes: unstable, learning, and overfitting}. We further characterize the minimal learnable parameters required for stable learning, uncover distinct convergence and pruning phases, and quantify the effect of reduced numeric precision on trainable parameters. Aligning with intuition, the results confirm that deeper architectures are more resilient to pruning than shallower ones, with parameter redundancy as high as $ 60\%$, and quantization impacts models with fewer learnable parameters more severely and has a larger effect on harder image datasets. These findings provide actionable guidance for selecting compact, stable models under pruning and low-precision constraints in image classification.
\end{abstract}
%(b) A  network pruning exploration to find minimal networks. (c) A network 8-bit quantization exploration is used to analyze performance deviation against 32-bit networks. Our systematic network design exploration includes a projection of the previous hidden layer to the next hidden layer with the same dimension (a square matrix), expanding dimension (low to high trapezoid shape matrix), and contracting dimension (high to low trapezoid shape matrix). Experiments were conducted on the most popular MNIST (as experimental protocols were extremely large and computationally demanding). Our results demonstrate that solving this image classification problem has three phases: non-optimal, sub-optimal, and optimal, and the phases correspond to the order magnitude increase of DNN learnable parameters $10^3$, $10^4$, and $10^5$. This performance is invariant across DNN architectures and the number of hidden layers used in DNNs. On pruning, the performance phases were clearly split on 60\%, and quantization was found to affect the model with a lower order of learning parameters. Thus, our results uncover the phases of DNN performances related to the variance in DNN architecture designs.

%Deep Neural Network, Minimal Neural Network, Network Pruning, Quantization-Aware Training, Sparsity

% %\begin{IEEEkeywords}
% Deep Neural Network, Minimal Neural Network, Network Pruning, Quantization-Aware Training, Sparsity
% % Giuseppe: why to reduce the number of keywords? ;-)-
% %\end{IEEEkeywords}

\section{Introduction}
Deep Neural Networks (DNNs) have achieved remarkable success in machine learning, particularly in image recognition~\cite{krizhevsky2012imagenet}. Beyond standalone multi-layer perceptrons (MLPs), dense layers are central components in modern architectures such as Convolutional Neural Networks (CNNs) and Transformers/Vision Transformers (ViTs)~\cite{he2024transformers}. Recent studies suggest that carefully designed MLPs can rival more complex models across a range of tasks~\cite{tong2024mlps,nie2021mlp}, underscoring the importance of understanding network compactness.

Despite this progress, discovering \emph{minimal} networks that solve a task efficiently remains challenging.%~\cite{pardalos2021black}. 
Minimal networks are crucial for scalable AI systems as they expose the essential degrees of freedom required for learning. Pruning methods show that large models contain substantial redundancy: simplified subnetworks can retain high accuracy with far fewer parameters~\cite{dekhovich2024pruning}. Related findings in Transformer-like architectures indicate that selectively removing components yields efficiency gains with minimal performance loss~\cite{he2024transformers}. In parallel, quantization reduces computational and memory costs by lowering arithmetic precision~\cite{hubara2016binarized,ma2024era}.

Understanding the characteristics of minimal DNNs for \emph{stable learning} calls for a systematic study linking architectural design to convergence dynamics, pruning behavior, and quantization robustness. In this paper, we conduct such a study focused \emph{exclusively on image datasets of increasing difficulties}. Since our computational overhead is extremely expensive due to architectural design variations, we selected the most popular benchmarks. We first perform a structured sweep over architectural \emph{shapes}—i.e., patterns of increasing/decreasing hidden-layer widths—then extend the analysis from MLP-based classifiers to CNNs and ViTs. For representative models, we examine (i) convergence profiles and their phases, (ii) sensitivity to pruning at different depths and parameter budgets, and (iii) the effects of reduced numeric precision on trainable parameters.

Our main contributions are:
\begin{itemize}
    \item A unified methodology for analyzing \textit{stable learning} in \textit{minimal neural networks} via joint examination of convergence, pruning, and quantization, applied across MLPs/DNNs, CNNs, and ViTs on image datasets.
    \item A systematic investigation of how architectural depth and hidden-layer \emph{shapes} influence regime-wise learning dynamics, pruning tolerance, and quantization robustness under varying problem complexities.
    \item Empirical characterization of the relationships among unstable/stable learning, learnable parameter counts, pruning rates, and precision reduction, offering guidance for selecting compact, resource-efficient models. % (e.g., deeper architectures exhibiting higher pruning resilience and substantial parameter redundancy).
\end{itemize}

The remainder of the paper is organized as follows. Sec.~\ref{sec:related_work} reviews related work. Sec.~\ref{sec:method} presents methodology. Sec.~\ref{sec:results} reports and analyzes experimental results for DNNs, CNNs, and ViTs. Sec.~\ref{sec:conclusion} concludes findings.

\section{Related Work} 
\label{sec:related_work}
Recent research underscores the rationale for seeking minimal networks by demonstrating that high performance can be achieved with fewer parameters. The lottery ticket hypothesis by Frankle and Carbin~\cite{frankle2018lottery} shows that sparse sub-networks, if properly initialized, can perform on par with their larger counterparts. Similarly, Neyshabur et al.'s work on over-parameterization and generalization~\cite{neyshabur2018understanding} shows that excessive parameter redundancy does not necessarily translate into improved network stability or efficiency. These findings align with our pursuit of profiling minimal networks and point to gaps in the research on systematically identifying and characterizing such architectures. A key factor in this exploration is the configuration of a network's hidden layers, which plays a significant role in learning and overall performance. Neural architecture search (NAS) techniques~\cite{zoph2017neural,liu2019darts} reveal that variations in layer configurations can fundamentally influence the network’s efficiency and generalization capability. Complementary insights from Naitzat et al.~\cite{naitzat2020topology} emphasize that the structural design of hidden layers affects how data representations are formed. Thus, investigating hidden layer combinations is crucial in understanding the intrinsic factors that can lead to minimal yet effective network architectures.

In studies of rational model minimization, pruning and quantization have emerged as key techniques for reducing model complexity. Molchanov et al.~\cite{molchanov2017variational} demonstrated that pruning could eliminate redundant parameters while maintaining performance, and practical low-precision methods show that constraining numerical precision to 8-bit can preserve accuracy while reducing cost~\cite{jacob2018quantization,rastegari2016xnor}. These studies give us insights into how to obtain the minimal network structure to explore the stable regimes of learning related to pruning and the impact of quantization on networks with different parameter orders. Kingma and Ba~\cite{kingma2014adam} demonstrated that Adam can substantially reduce the loss within a relatively small number of epochs, indicating that even a short training period (on the order of 20 epochs) may reveal significant convergence trends. In a related study, Smith~\cite{smith2017cyclical} introduced cyclical learning rates, showing that performing a learning rate range test over approximately $10--20$ epochs is sufficient to capture the overall loss behavior and to identify an optimal learning rate range. Building on these foundations, our work integrates convergence-phase analysis, pruning phases, and quantization into basic networks. We conduct extensive experiments across multiple \emph{image} datasets under a unified experimental configuration and extend them to MLPs/DNNs, CNNs, and ViTs to explore how architectural decisions and parameter-level configurations affect network convergence and performance, thereby uncovering deeper insights into the design of minimal yet robust models.

\section{Methodology}\label{sec:method}

\subsection{Systematic Characterization}
\begin{figure}
    \centering
    \includegraphics[width=0.98\linewidth]{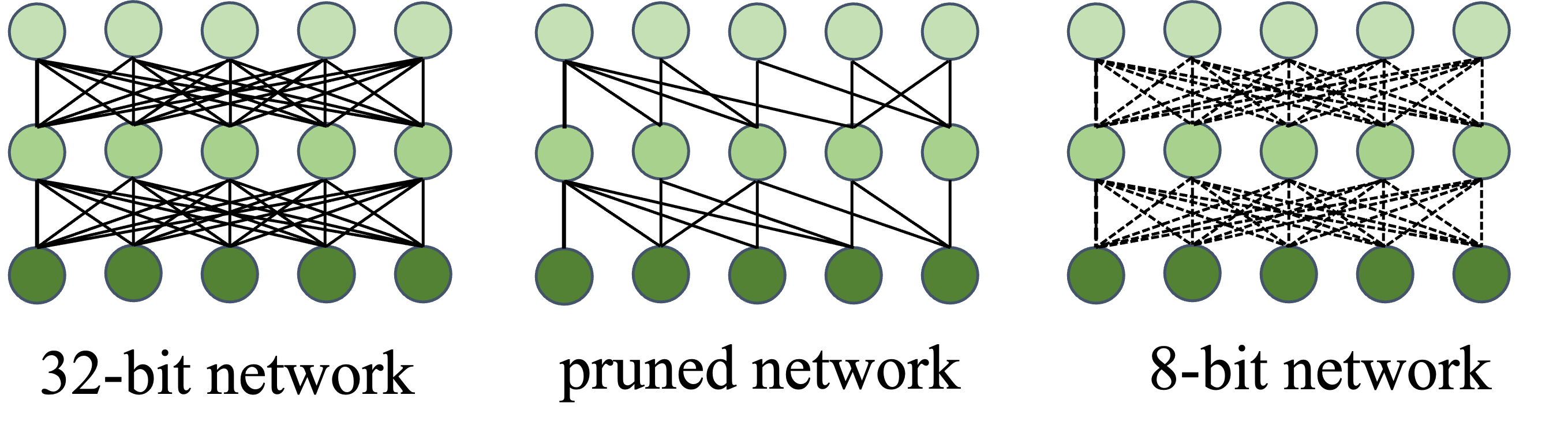}
    \caption{1-layer whole network, pruned network, and quantized network (see Sec.~\ref{sec:method} for details).}
    \label{fig:example_net}
\end{figure}

We employ a systematic architecture design to analyze three key phases--learning, pruning, and quantization--to yield a minimal neural network architecture (see an example 1-(hidden)layer network in Fig.~\ref{fig:example_net}). Our approaches enable a systematic evaluation of the topological (network configurations) and parametric factors that influence learning dynamics and network performance.

\begin{figure*}[h!]
    \setlength{\tabcolsep}{1pt}
    \centering
     \includegraphics[width=\textwidth]{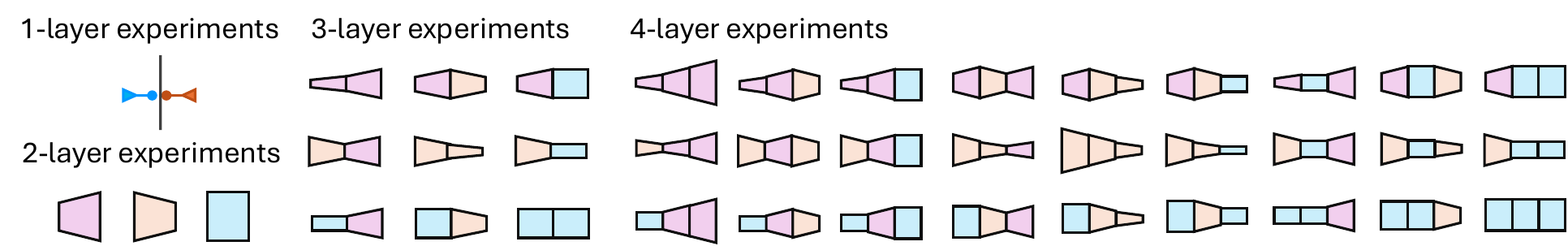} \\
     (a)\\
    \begin{tabular}{ccccc}
         & {\sc 1layer-arch} & {\sc 2layer-arch} & {\sc 3layer-arch} & {\sc 4layer-arch} \\
        \rotatebox[origin=c]{90}{{\sc MNIST}}
            & \raisebox{-0.5\height}{\includegraphics[width=0.24\textwidth]{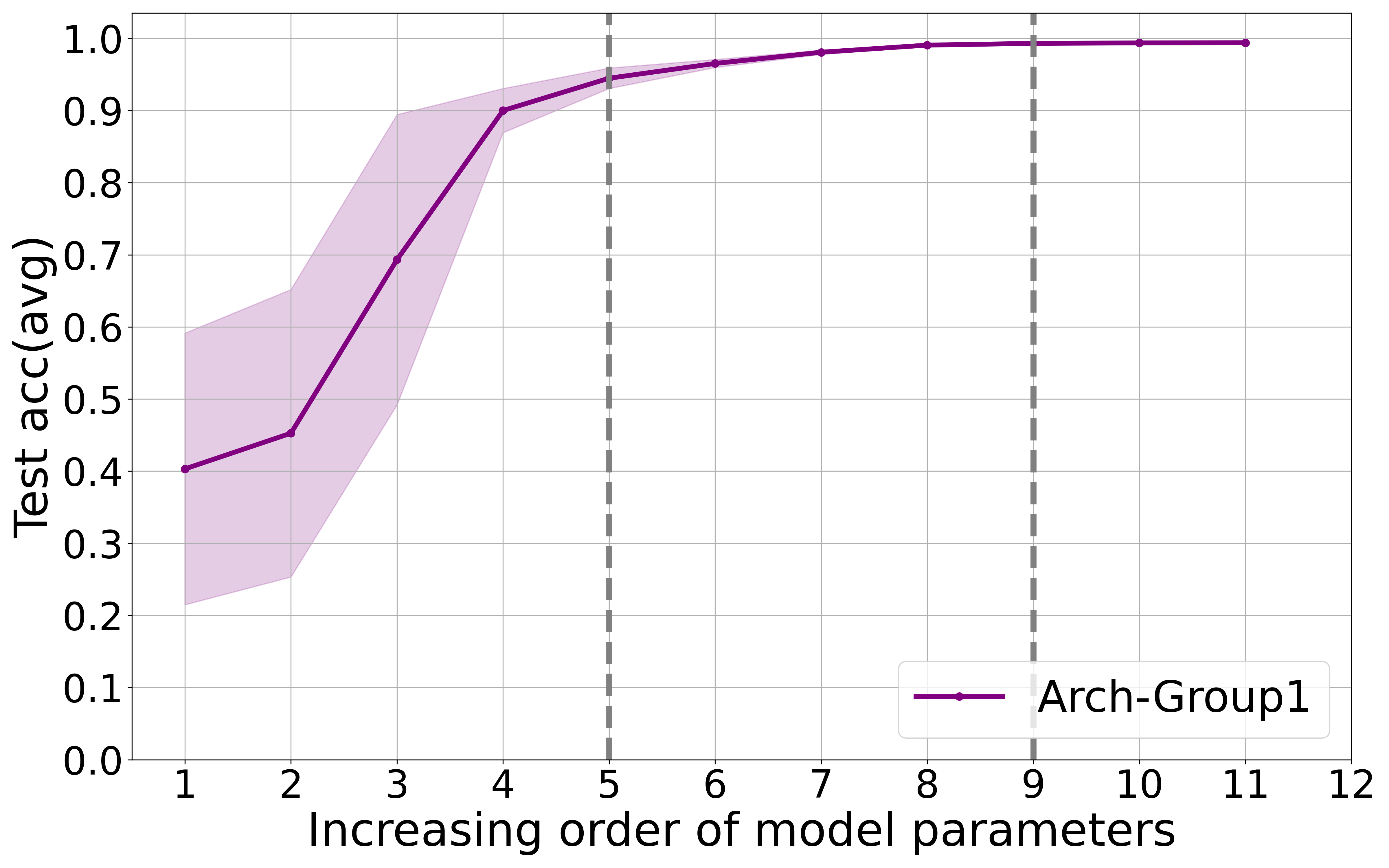}}
            & \raisebox{-0.5\height}{\includegraphics[width=0.24\textwidth]{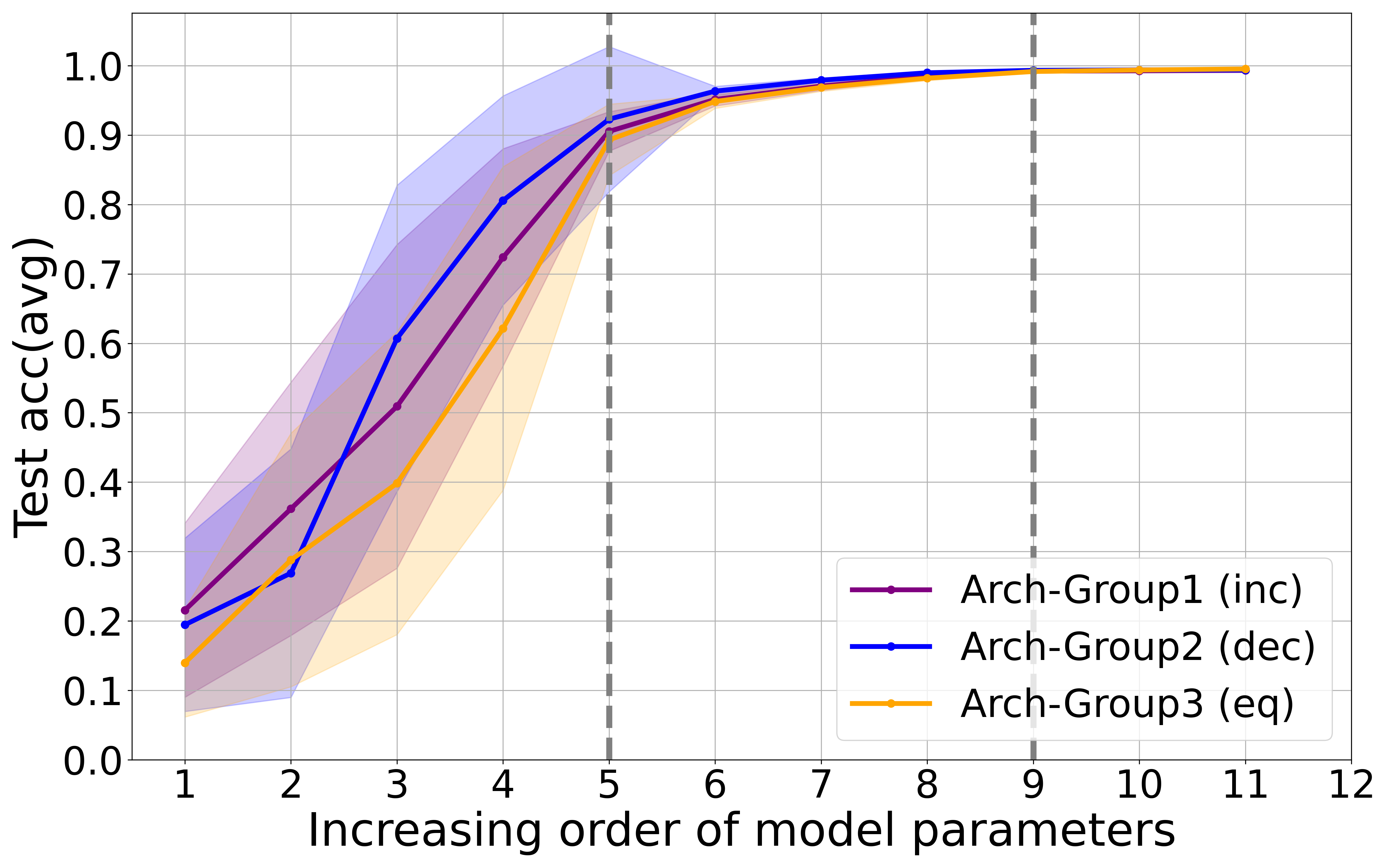}}
            & \raisebox{-0.5\height}{\includegraphics[width=0.24\textwidth]{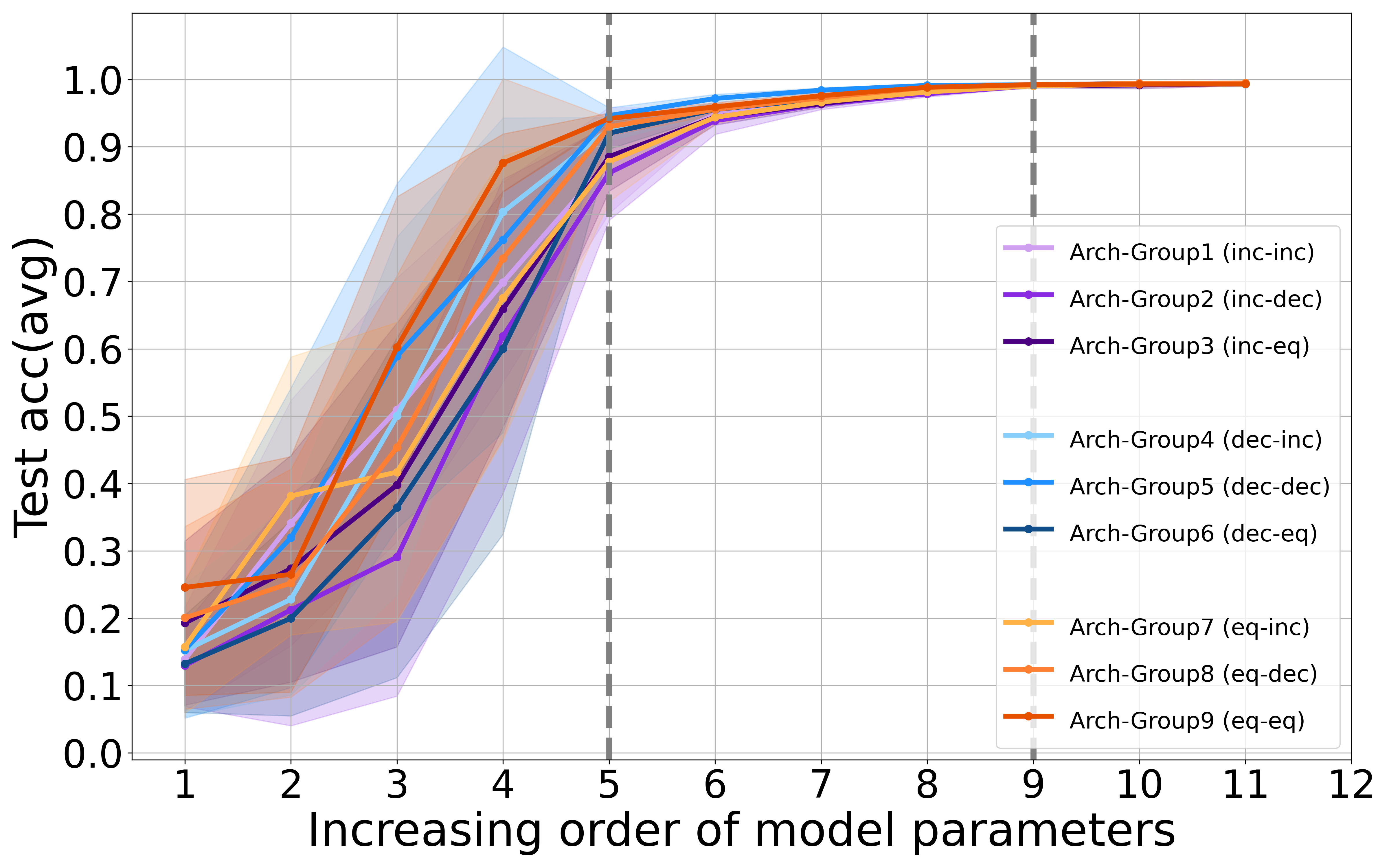}}
            & \raisebox{-0.5\height}{\includegraphics[width=0.24\textwidth]{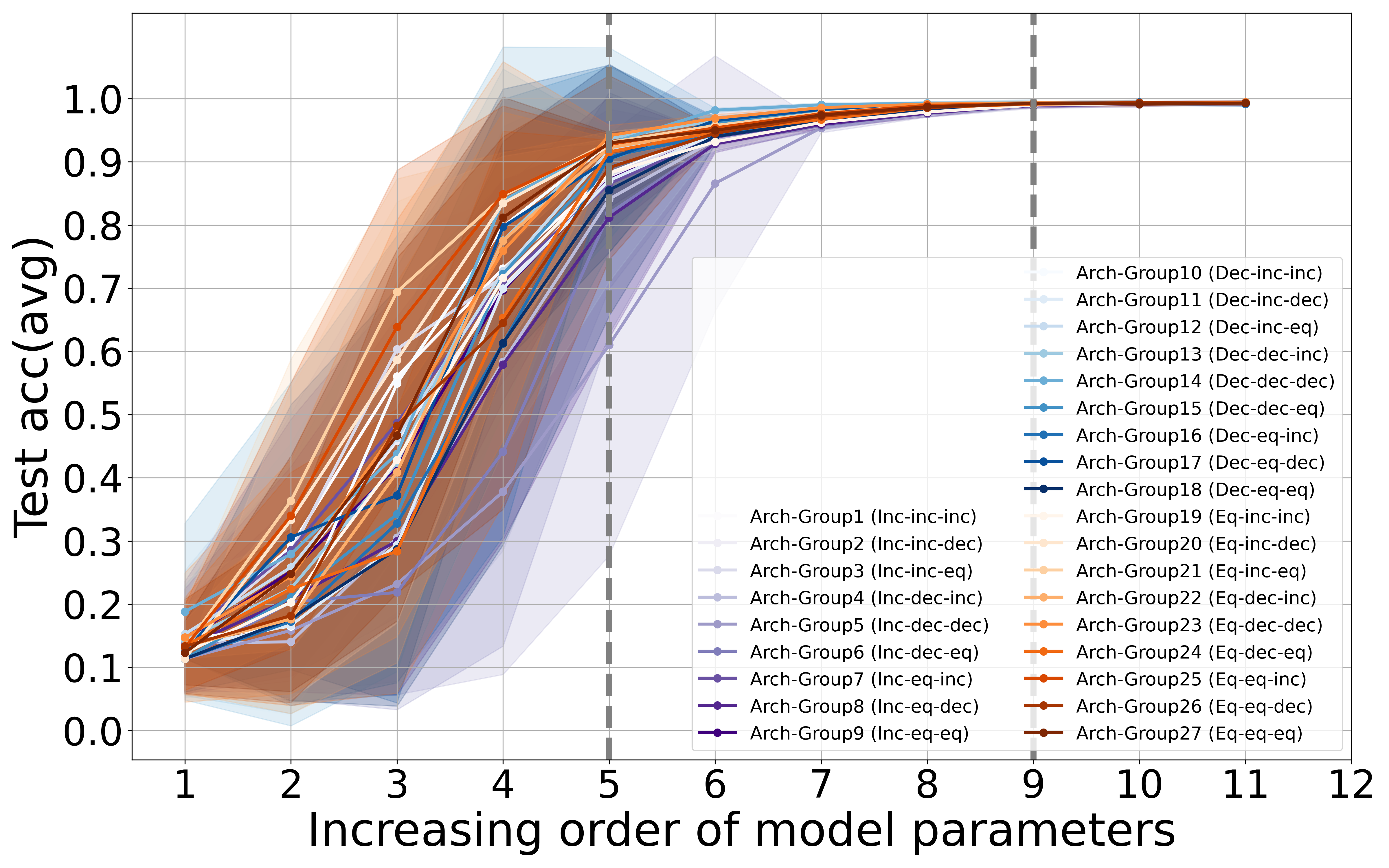}}\\
        \rotatebox[origin=c]{90}{{\sc FASHION MNIST}} 
            & \raisebox{-0.5\height}{\includegraphics[width=0.24\textwidth]{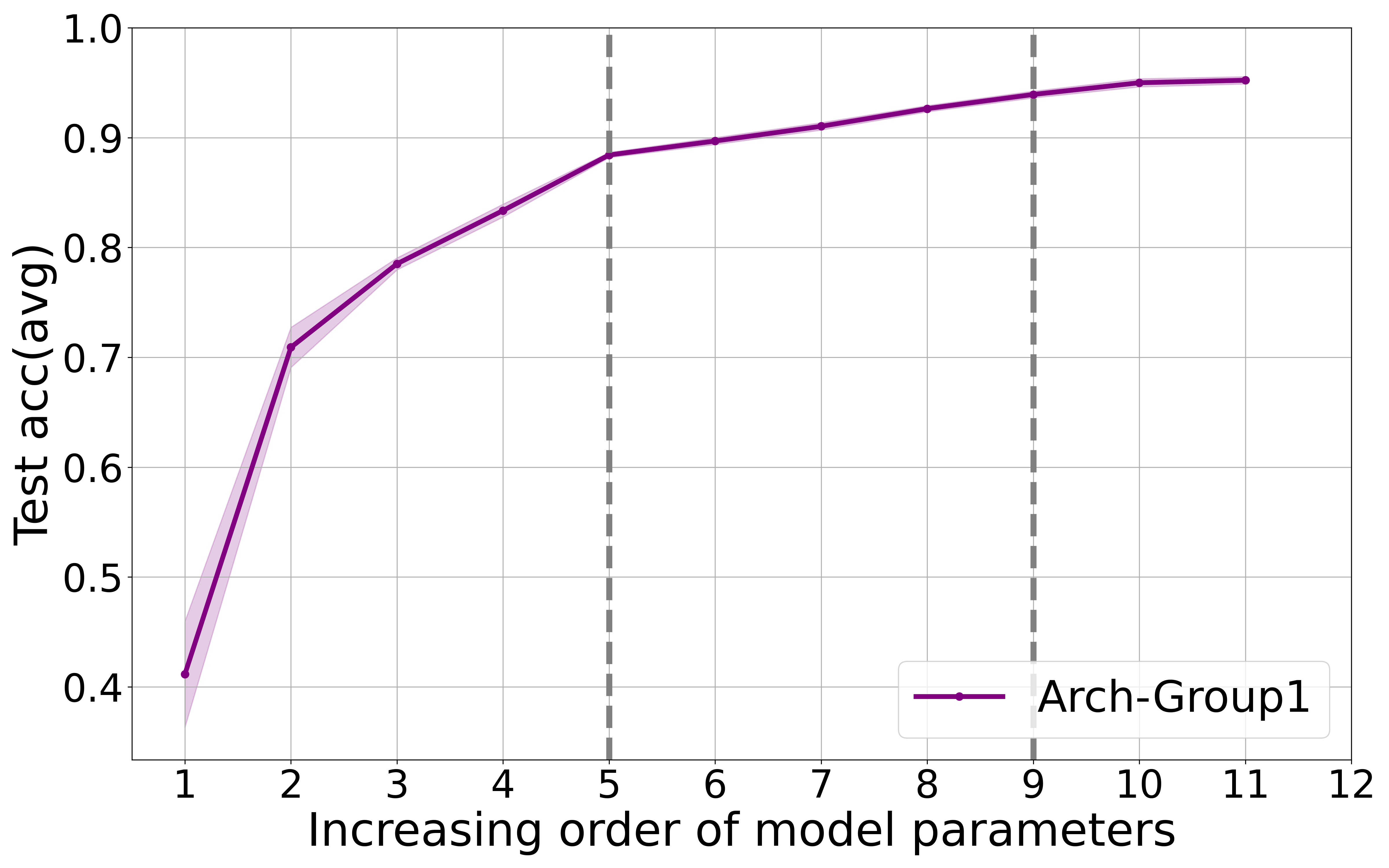}}
            & \raisebox{-0.5\height}{\includegraphics[width=0.24\textwidth]{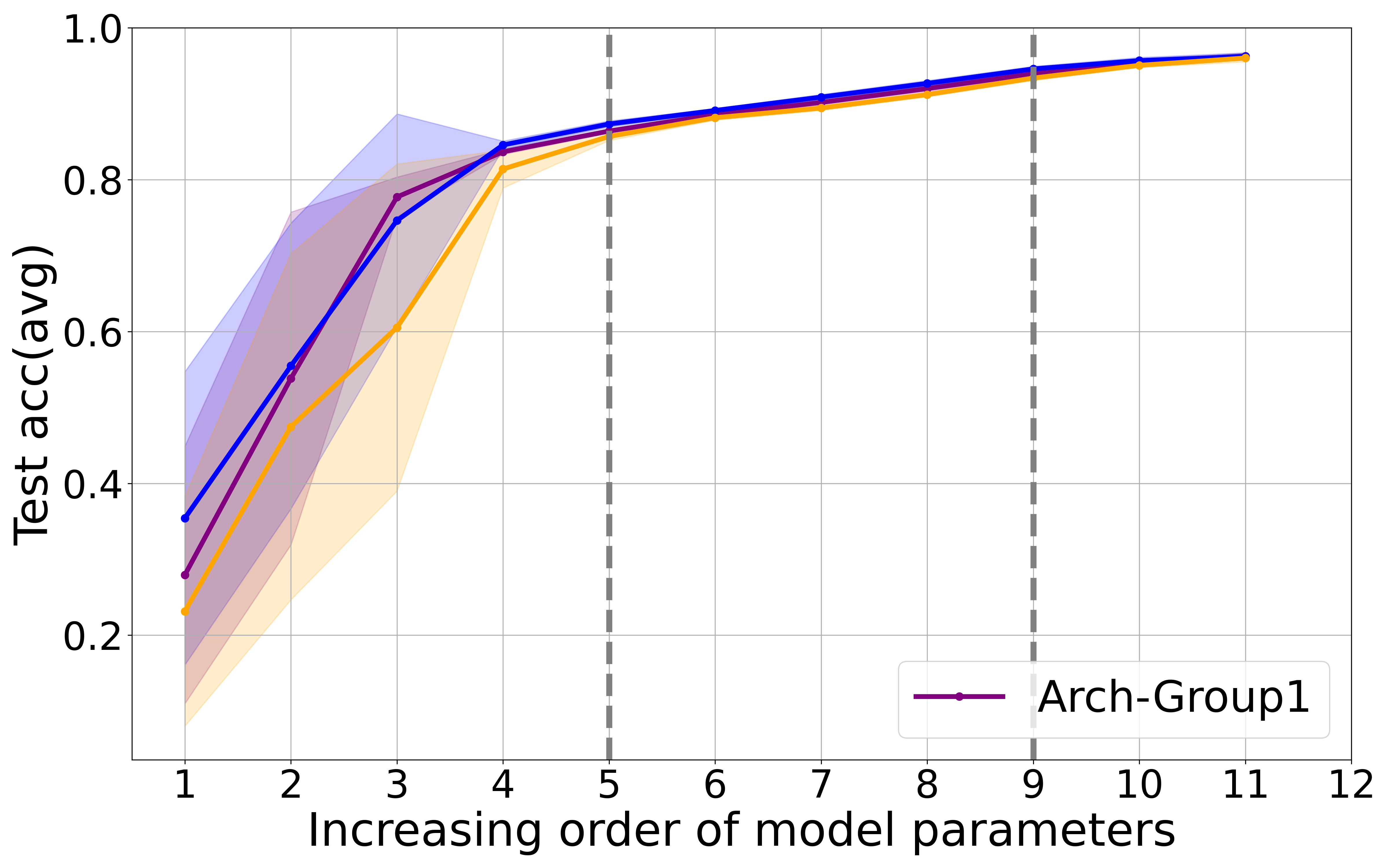}}
            & \raisebox{-0.5\height}{\includegraphics[width=0.24\textwidth]{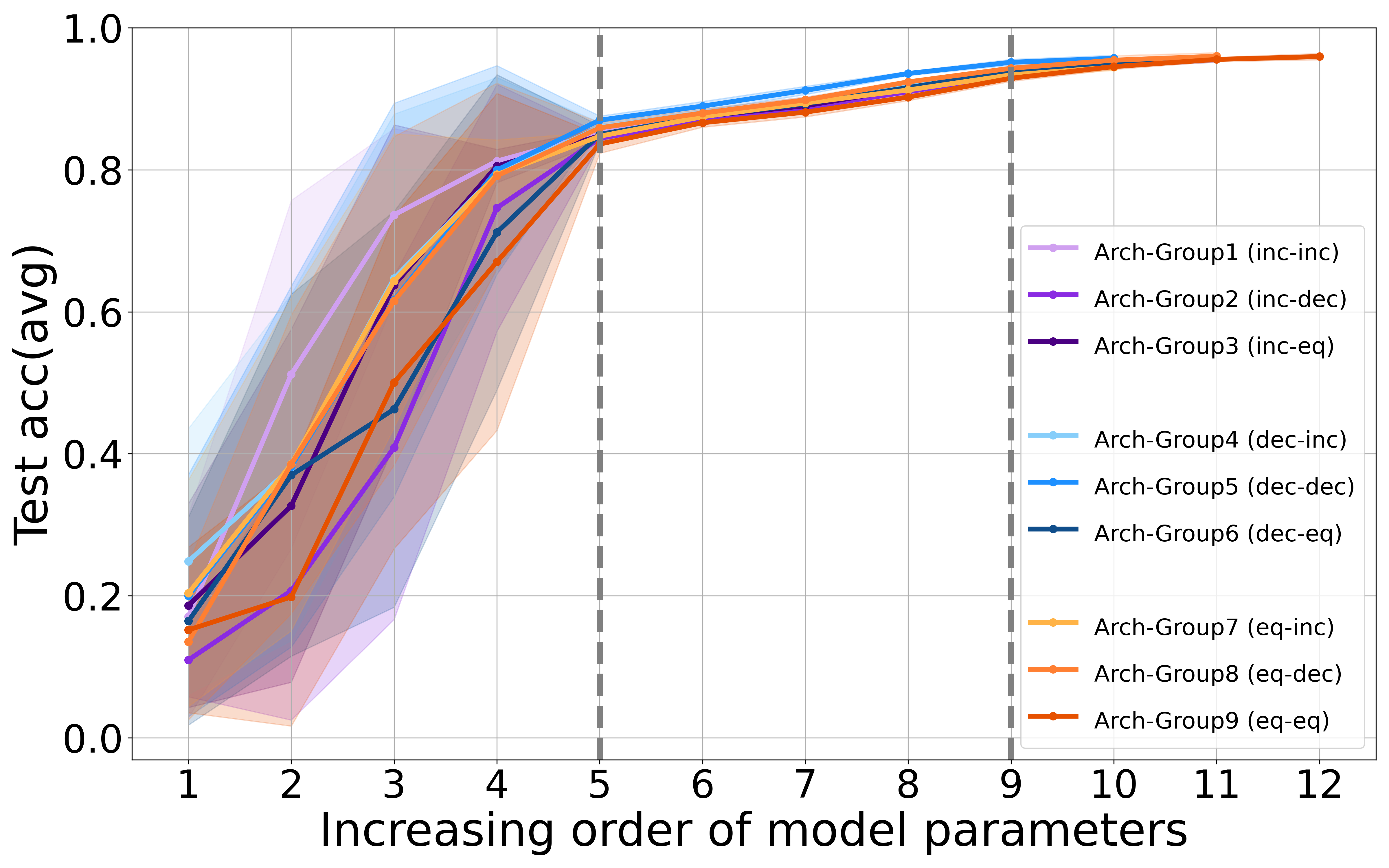}}
            & \raisebox{-0.5\height}{\includegraphics[width=0.24\textwidth]{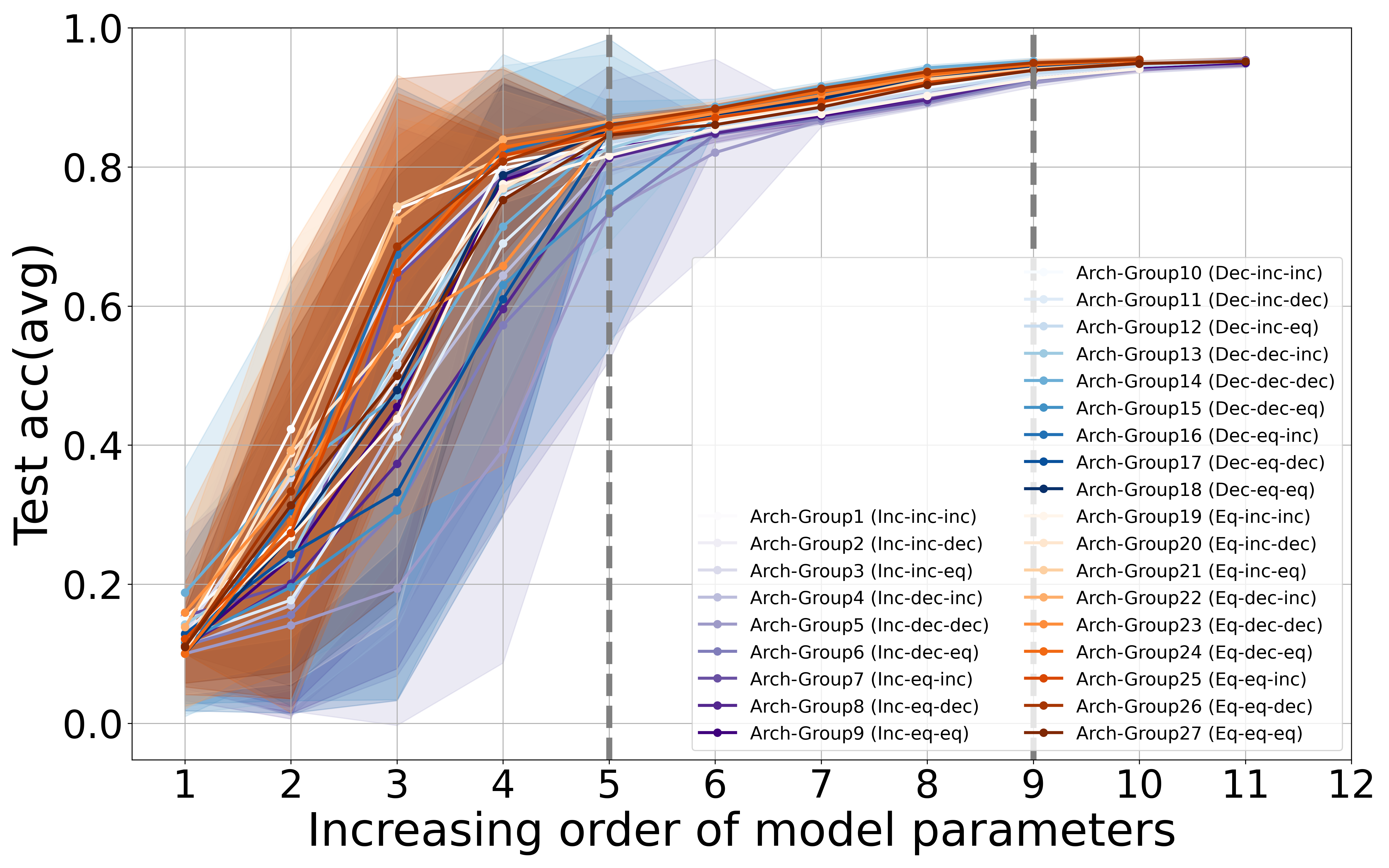}} \\
        \rotatebox[origin=c]{90}{{\sc CIFAR-10}} 
            & \raisebox{-0.5\height}{\includegraphics[width=0.24\textwidth]{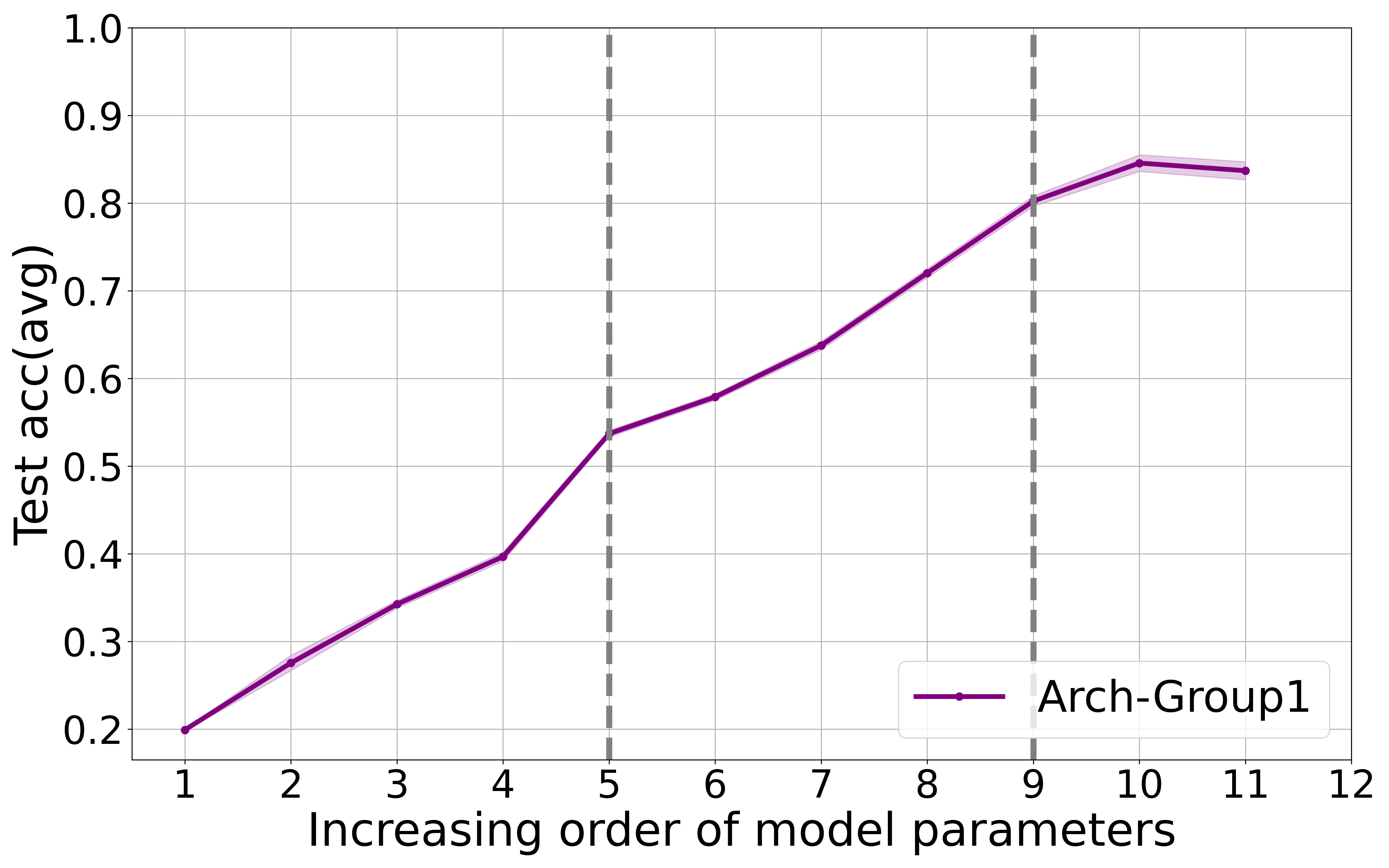}}
            & \raisebox{-0.5\height}{\includegraphics[width=0.24\textwidth]{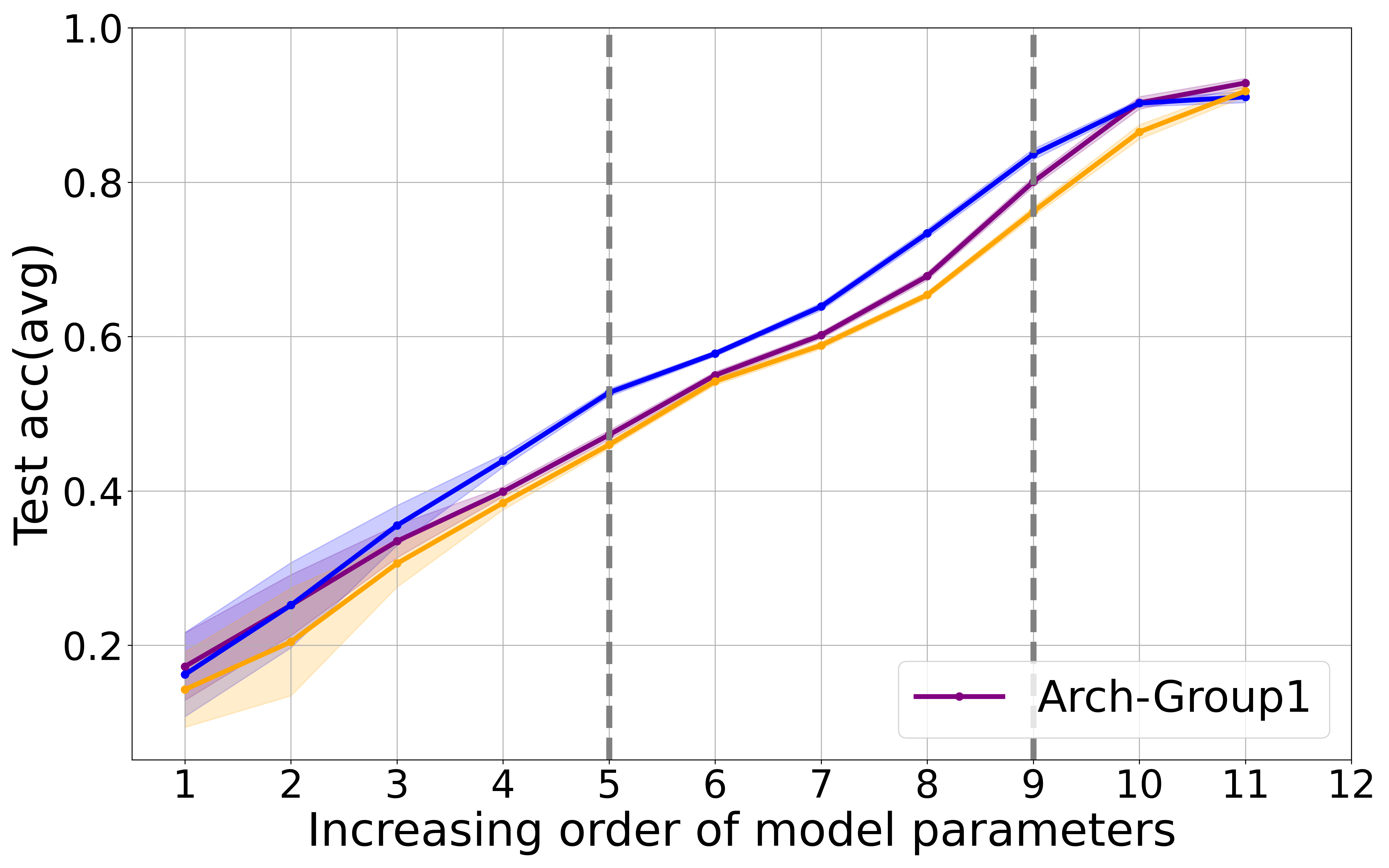}}
            & \raisebox{-0.5\height}{\includegraphics[width=0.24\textwidth]{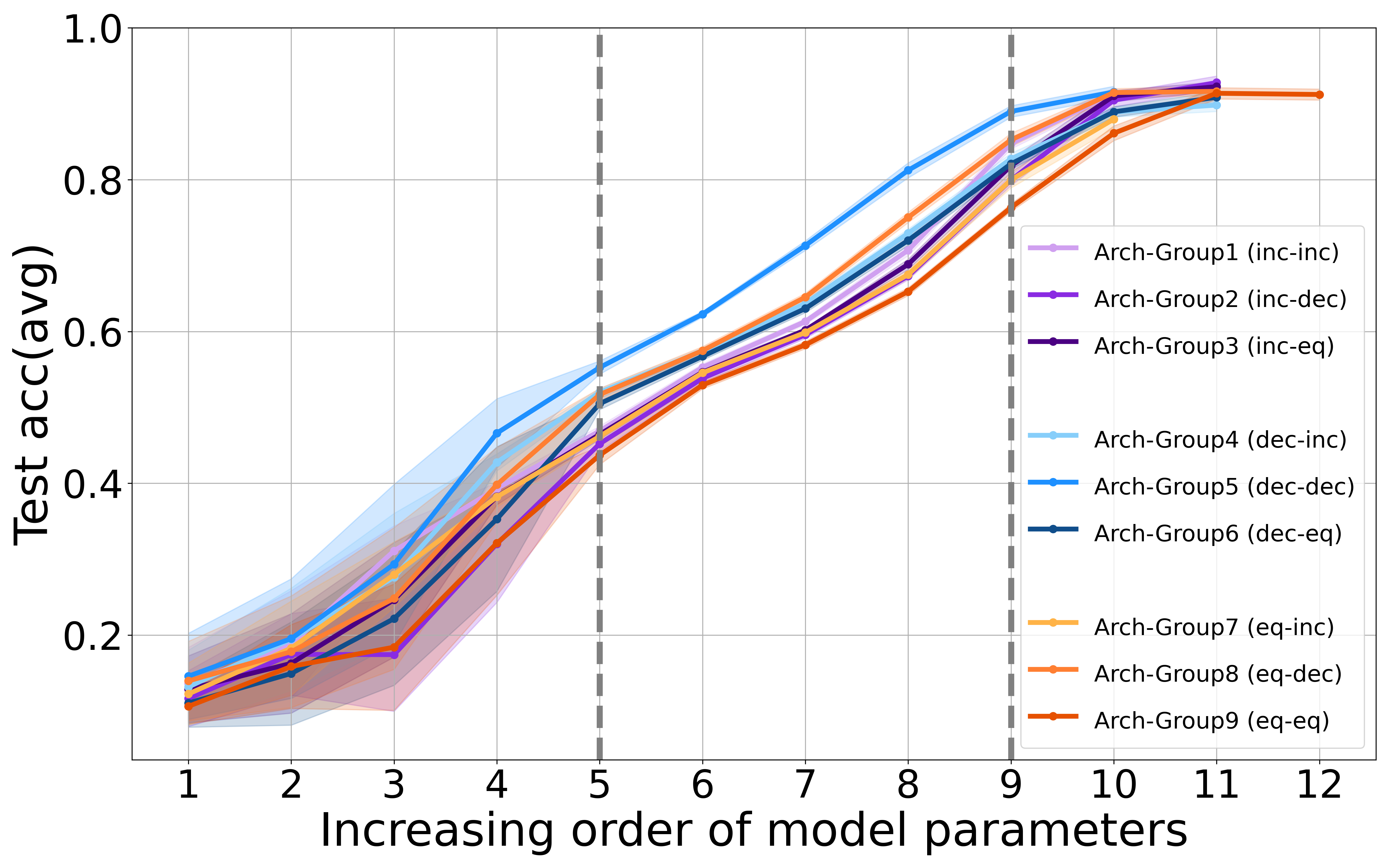}}
            & \raisebox{-0.5\height}{\includegraphics[width=0.24\textwidth]{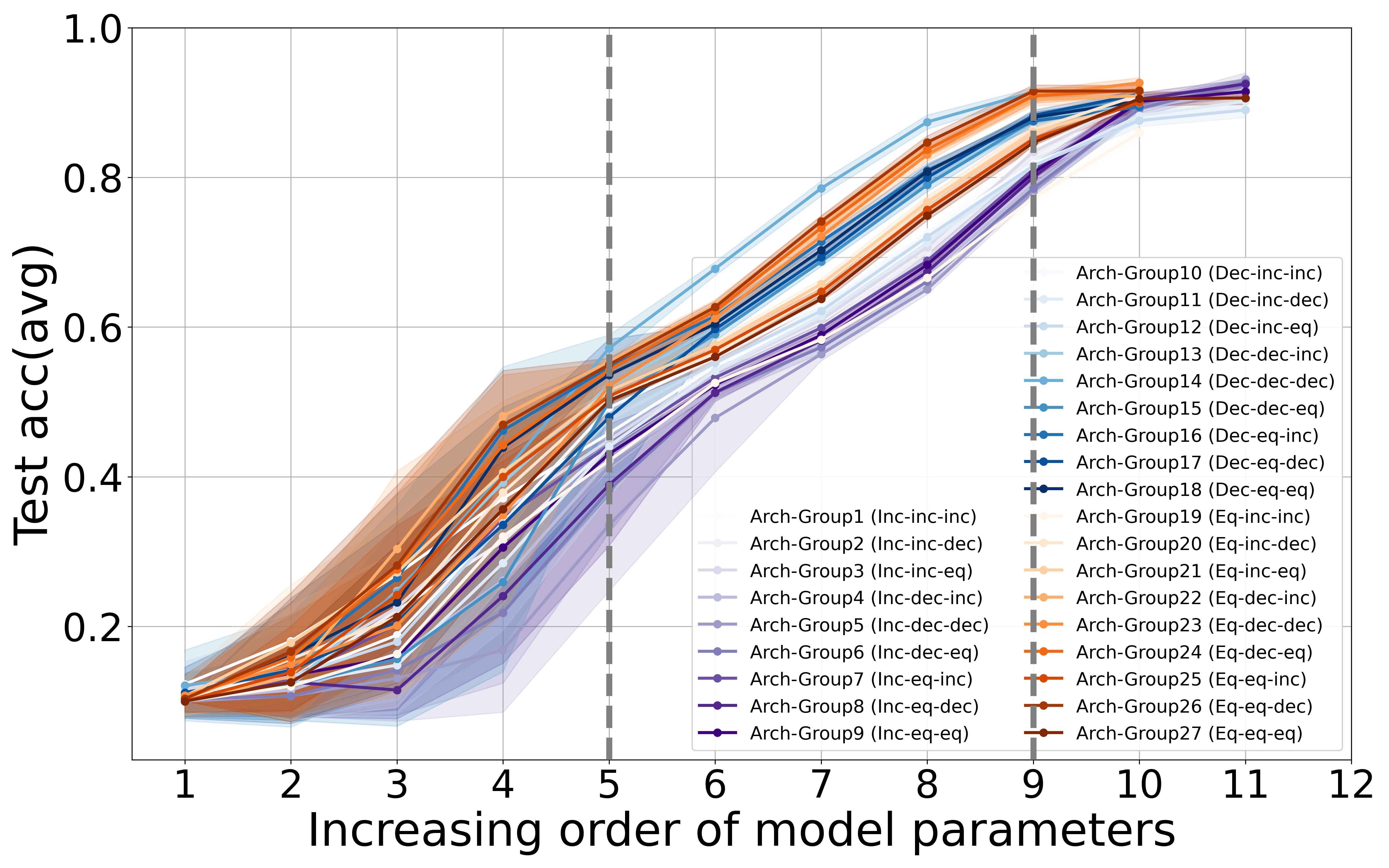}}
    \end{tabular}
    \newline
    (b)
    \caption{Network architecture design and performance. (a) The number indicates the number of hidden layers. The weight matrix shapes $\vartriangleleft$, $\vartriangleright$, and $\square$ stand for an increase, decrease, and an equal number of nodes in the consecutive layers, and $\mid$ represents a single hidden layer. (b) Left to right: convergence performance of 1-layer to 4-layer networks. The vertical dotted lines separate different ranges of the total number of learnable parameters.}
    \label{fig:alllayershape}
\end{figure*}

\paragraph{Convergence phases for stable learning}  
Neural network convergence has been a long-standing topic, with theoretical results demonstrating the approximation capabilities of MLPs/DNNs~\cite{hornik1989multilayer}. We therefore analyze the evolution of network empirical performance in relation to changes in network parameters and topologies. By monitoring the model's test accuracy across different parameter phases, we identify critical periods that significantly contribute to convergence and generalization. Our analysis of convergence profiles is computationally intensive. Thus, we divided it into two phases:

\textit{First phase analysis.} In the initial phase, we varied the depth of neural network layers and the topologies (the shapes of the weight matrices between hidden layers), including increases, decreases, and equal numbers of nodes in consecutive layers. This gives us a comprehensive set of network topologies (see Fig.~\ref{fig:alllayershape}(a)). For each network topology in Fig.~\ref{fig:alllayershape}(a), we varied network architecture from a bare minimum of one hidden node to 2, 3, 5, 10, 20, 30, 50, 100, 200, 500, and 1000 nodes at the respective layers of network architecture with 1-layer to 4-layer. This means that for a 1-layer experiment, there were 12 sets. For 2-layer, 3-layer, and 4-layer architectures, there were 3 (topologies) × 12 (network sizes), 9 (topologies) × 12 (network sizes), and 27 (topologies) × 12 (network sizes) experiments, respectively. We repeated each experiment 30 times for robust results.

\textit{Second phase analysis.} In the second phase, we narrow down the computational overhead based on the observation of the initial phase (see Fig.~\ref{fig:alllayershape}(b)) that the topological variations have an invariant influence on convergence performance for larger parameter models. That is, \textit{the size of the learnable parameters has a dominant influence on the convergence profile compared to topological variations}. Thus, for the second phase, we select networks with equal numbers of nodes in successive layers and vary the number of nodes to 1, 2, 3, 5, 10, 20, 30, 50, 100, 200, 500, and 1000 for MLPs/DNNs. In the second phase, we extend the study to CNNs and ViTs with several architectural variations. 
As shown in Fig.~\ref{fig:alllayershape}(b), the convergence curves for varying layer shapes largely overlap once the network size exceeds a certain parameter count (e.g., after parameter order $6-7$), indicating that model size, not shape, is the dominant factor for stable performance.
This analysis allows us to uncover the influence of training dynamics on model stability and performance.

\paragraph{Pruning phase of stable learning}  
Studies have shown that simple magnitude-based pruning (L1 pruning) is as effective as complex pruning methods~\cite{redman2021operator}. However, there is a lack of theoretical underpinnings for the success of magnitude pruning, and it remains an outstanding open question in the field~\cite{redman2021operator}. Some theoretical results classify pruning into mild pruning and over-pruning and offer a theoretical perspective on pruning~\cite{yang2025random}. Our study provides an empirical perspective into L1 pruning~\cite{han2015deep} due to its computational simplicity and effectiveness. L1 pruning uses the absolute value of each weight as an indicator of its contribution, providing a uniform criterion for evaluating weight importance without favoring any specific layer or connection. This `fair' approach allows all weights to be assessed on the same scale, which aligns well with our focus on analyzing {\it phase transitions} in network performance as a function of pruning severity (pruning rate). We conducted a pruning analysis by varying the pruning rate from 10\% to 100\% on the best architecture that offers stable learning in the second phase of convergence analysis.

\paragraph{Quantization phase stable learning}  
Quantization is applied during training to simulate lower-precision \textit{weights} and \textit{activations}, thereby enabling efficient deployment on resource-constrained devices. Several studies demonstrate the difficulty of training with low-precision arithmetic and provide theoretical analyses of behavior and the reasons for the success or failure of various quantization methods~\cite{li2017training,jacob2018quantization}. In our work, we present empirical profiles of success and failure cases by adopting Quantization-Aware Training (QAT)~\cite{jacob2018quantization}, which enables the network to adapt to quantization effects during training, resulting in more robust performance than post-training quantization methods. We use 8-bit integer quantization (W8A8) with PyTorch’s default QAT configuration for the QNNPACK backend: weights are quantized to signed 8-bit integers (\texttt{qint8}) using per-tensor \emph{symmetric} uniform quantization, while activations are quantized to unsigned 8-bit integers (\texttt{quint8}) using per-tensor \emph{asymmetric} (affine) quantization. The corresponding quantization parameters (scale and zero-point) are estimated by a MovingAverageMinMax observer that maintains running estimates of the activation and weight ranges during training, effectively using a max-abs criterion for symmetric weight quantization and min–max statistics for asymmetric activation quantization. %QAT quantizes weights and activations during the forward pass and uses a straight-through estimator during backpropagation, helping maintain gradient flow and mitigating accuracy loss. We, therefore, compare the full-precision model's accuracy (unquantized) with the quantized model's accuracy to evaluate the loss induced by quantization. We performed experiments on the best architectures selected from the second phase.

\subsection{CNN and ViT architectures}
\paragraph{CNN network}
CNN network maps an input image to class probabilities through a set of feature extractors followed by a linear classifier. The feature extractor consists of a single spatial convolution with a kernel size of \(3\times3\), unit stride, and no padding, which emphasizes local edge and texture patterns while preserving most of the spatial extent. A rectified linear (ReLU) activation introduces nonlinearity at the feature level. The resulting activation tensor is flattened into a vector and passed to a fully connected layer that produces class logits; during training and evaluation, probabilities are obtained with a softmax. This design deliberately concentrates representational capacity in the number of output feature channels of the first convolution, enabling precise control of the total learnable parameters while keeping depth and spatial operations fixed. The architecture is used across image datasets, including single- and multi-channel inputs, with the classifier input dimension adjusted to the post-convolution spatial size and channel counts. 

\paragraph{ViT network}
We follow~\cite{dosovitskiy2020image} for the ViT network implementation, where each image is partitioned into a uniform grid of non-overlapping patches with exact divisibility along height and width. Every patch is flattened and linearly projected to a fixed-width embedding, producing a sequence of patch tokens. A learnable class token is added to this sequence, and fixed positional embeddings are added to encode spatial order. A stack of transformer encoder processes the token sequence; each block applies multi-head self-attention, followed by a position-wise feed-forward network, with residual connections and layer normalization applied to both sublayers. The representation associated with the class token at the encoder's output is mapped to class logits by a linear layer, and softmax yields class probabilities. The \textit{hidden dimension} primarily controls model capacity. In contrast, the patch grid, the number of encoder blocks, and the number of attention heads are kept constant to ensure consistent comparisons of convergence behavior, pruning tolerance, and quantization effects.

\subsection{Experimental Setup}\label{sec:experiment}
We conducted experiments on widely used image classification benchmarks (MNIST, Fashion-MNIST (FMNIST), and CIFAR-10). The increasing order of complexity of the image datasets is as follows: MNIST (gray images of 10 classes of handwritten characters), FMNIST (gray images of 10 classes of clothing), and CIFAR-10 (3-channel images of 10 classes of natural objects such as cars, planes, and horses).
\paragraph{DNN-specific configuration}  
%\textit{Training hyperparameter setting.}
The hidden layers use ReLU activation, and all network weights are initialized from a zero-mean normal distribution. Unless otherwise stated, training uses a batch size of $100$, a learning rate of $0.001$ with the Adam optimizer, $20$ training epochs, and cross-entropy loss. Each experiment is repeated $30$ times to ensure statistical reliability. 10\% of the training split is used for early stopping. Inputs are normalized channel-wise with mean $0.5$ and standard deviation $0.5$. 

\paragraph{CNN-specific configuration.}
The convolutional classifier employs a single spatial convolution with a $3\times3$ kernel, unit stride, and no padding, followed by a ReLU activation and a linear classifier. Model capacity is controlled by varying the number of output feature channels in the convolutional layer between $1$ and $32$, while keeping all other architectural choices fixed; the classifier's input dimension is adjusted to the resulting spatial resolution and channel count. Convolutional kernels are initialized with Kaiming He normal initialization, scaled by the receptive field and the number of output channels; linear layers are initialized with a small-variance normal distribution and zero biases. For datasets with different input modalities, the number of input channels in the first convolution matches the dataset, and the rest of the network remains unchanged.

\paragraph{ViT-specific configuration.}
The Vision Transformer (ViT) partitions each image into a uniform $7\times7$ grid of non-overlapping patches. ViT projects each flattened patch to a fixed-width embedding (\textit{hidden dimension}) and adds an extra learnable class token. The ViT encoder contains two multi-head self-attention (MHSA) blocks and a feed-forward MLP. The MHSA block, MLP, and Norm/layers use the same \textit{hidden dimension} as the size of the hidden dimension of the patch embedding layer. The class token representation at the encoder output is mapped to class logits via a linear layer. Model capacity is controlled by sweeping over the hidden dimensions between $1$ and $16$, while keeping the patch grid, the number of encoder blocks, and the number of attention heads fixed. The settings for optimizer, batch size, number of epochs, loss function, and hardware environment are identical to those used for the DNNs and CNNs.

All experiments are executed on a Windows x86 system with an Intel i9-10900X CPU and $32$\,GB RAM. This common setup enables controlled comparisons of learning dynamics, pruning behavior, and quantization effects. Unlike MLP/DNN experiments, which were repeated 30 times due to computational overhead, CNN and ViT experiments, due to their high computational cost, were repeated 10 times.

\section{Results and Analysis}\label{sec:results}
%The results of our experiment are shown in Fig.~\ref{fig:results}.
The results from the second phase of analysis, where architectures are a square shape matrix (Fig.~\ref{fig:alllayershape}), are shown in Figs.~\ref{fig:results_dnn}, ~\ref{fig:results_cnn}, and ~\ref{fig:results_vit} and Table~\ref{tab:stability_pruning_quant}. The results of convergence, pruning, and quantization are discussed as follows:

\begin{figure*}[h!]
    %\footnotesize
    \setlength{\tabcolsep}{5pt}
    \centering
    \begin{tabular}{cccc}
    \centering
         & {\sc MNIST} & {\sc Fashion MNIST} & {\sc CIFAR-10} \\ %& {\sc CIFAR-100} \\
        \rotatebox[origin=c]{90}{{\sc Convergence}}
            & \raisebox{-0.5\height}{\includegraphics[width=0.28\textwidth]{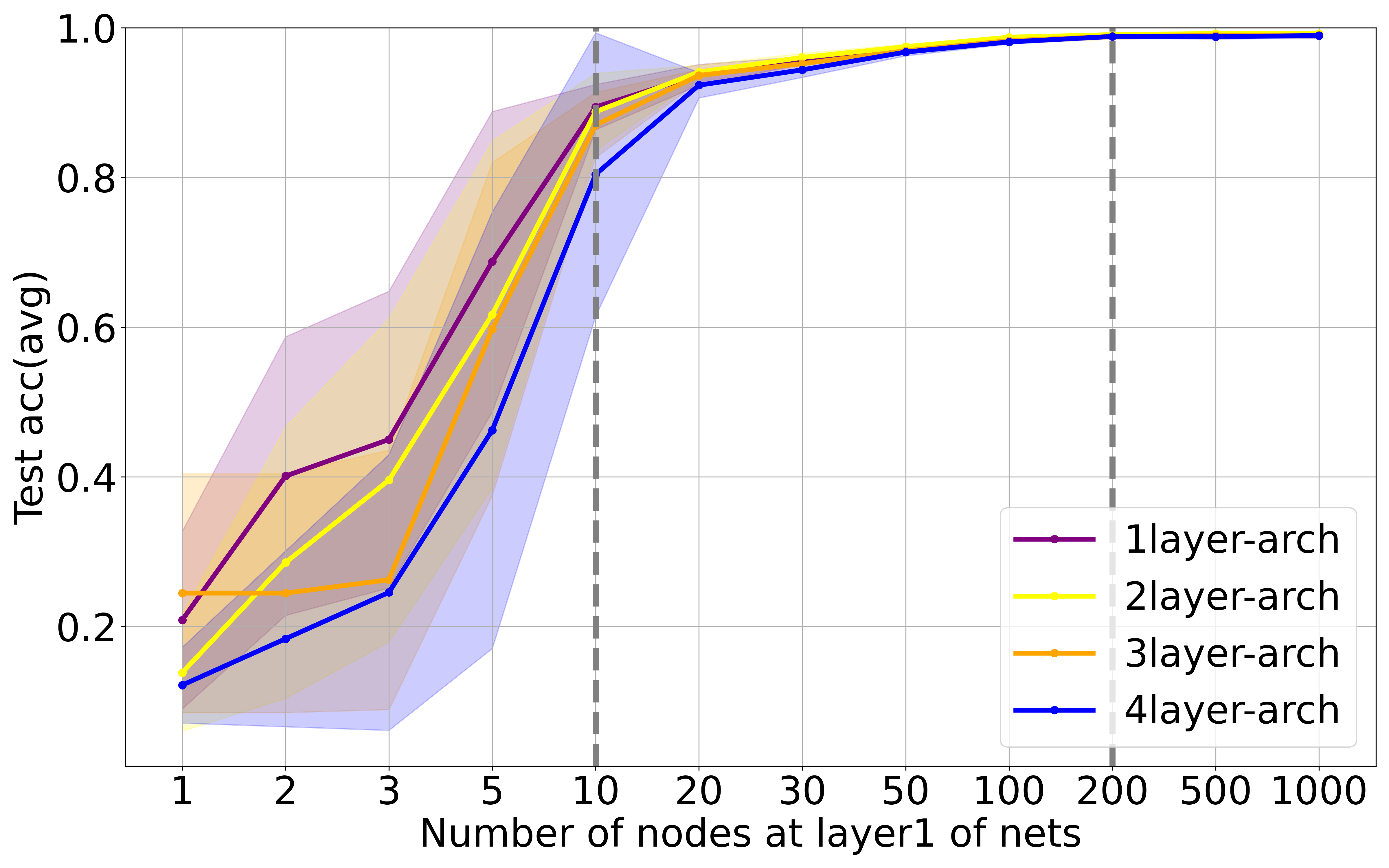}} ~\quad % width=0.24 
            & \raisebox{-0.5\height}{\includegraphics[width=0.28\textwidth] {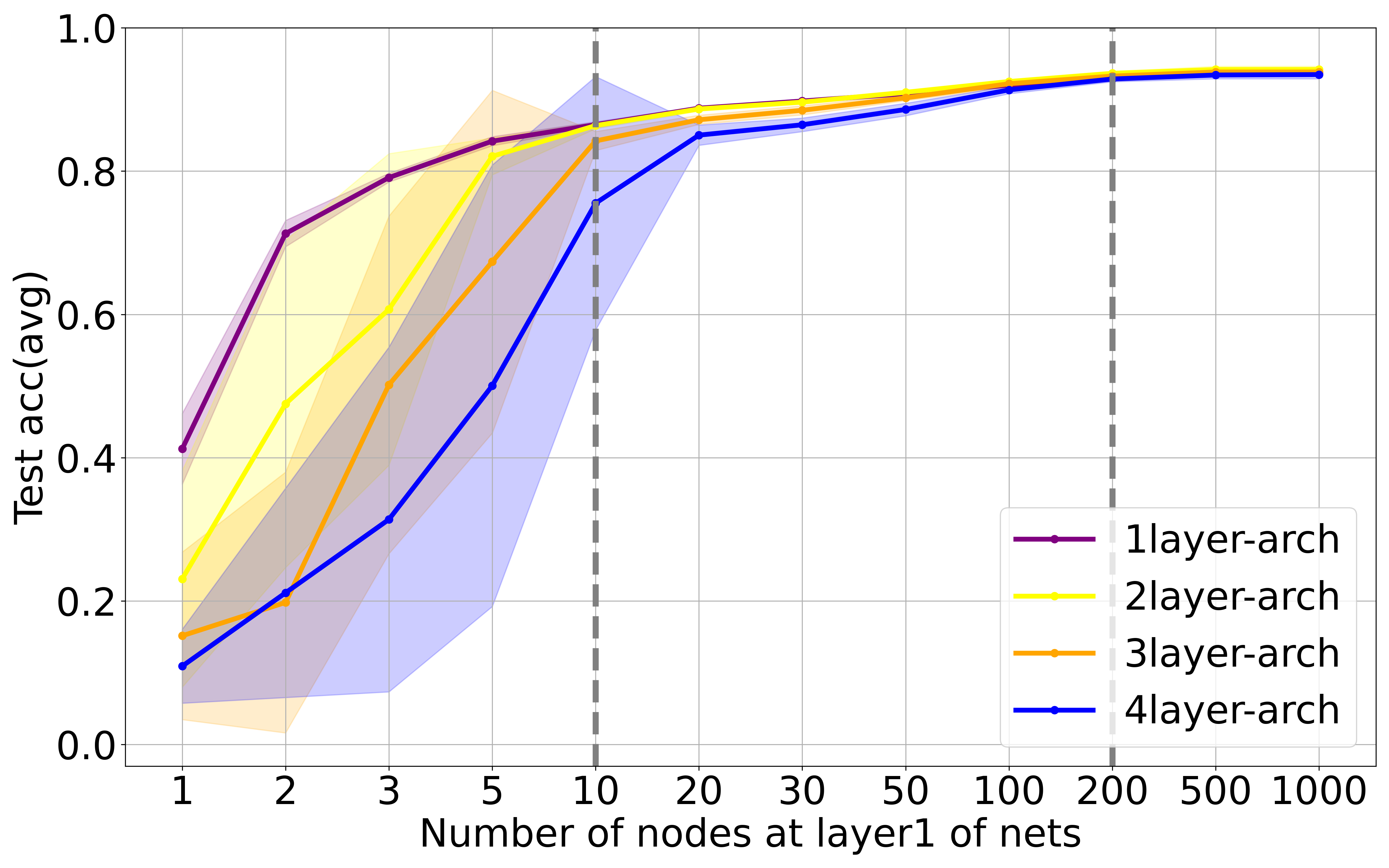}} ~\quad
            %\scriptsize (a) & \scriptsize (b) 
            & \raisebox{-0.5\height}{\includegraphics[width=0.28\textwidth]{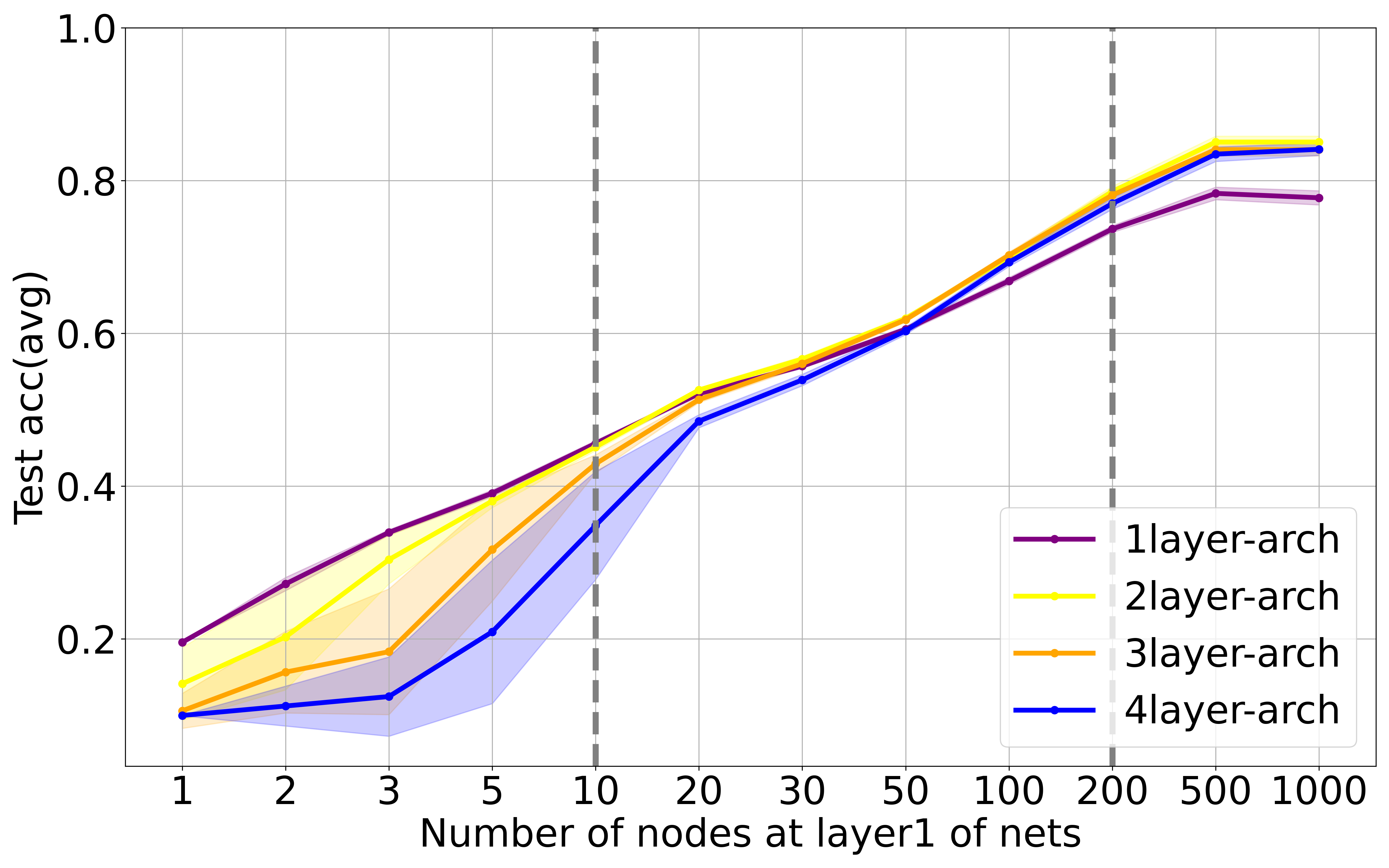}}\\
            %& \raisebox{-0.5\height}{\includegraphics[width=0.24\textwidth]{dnn/phase_mlp_cifar100_alllayer_nodes_interpolated.png}}\\
        
        \rotatebox[origin=c]{90}{{\sc Pruning}} 
            & \raisebox{-0.5\height}{\includegraphics[width=0.28\textwidth]{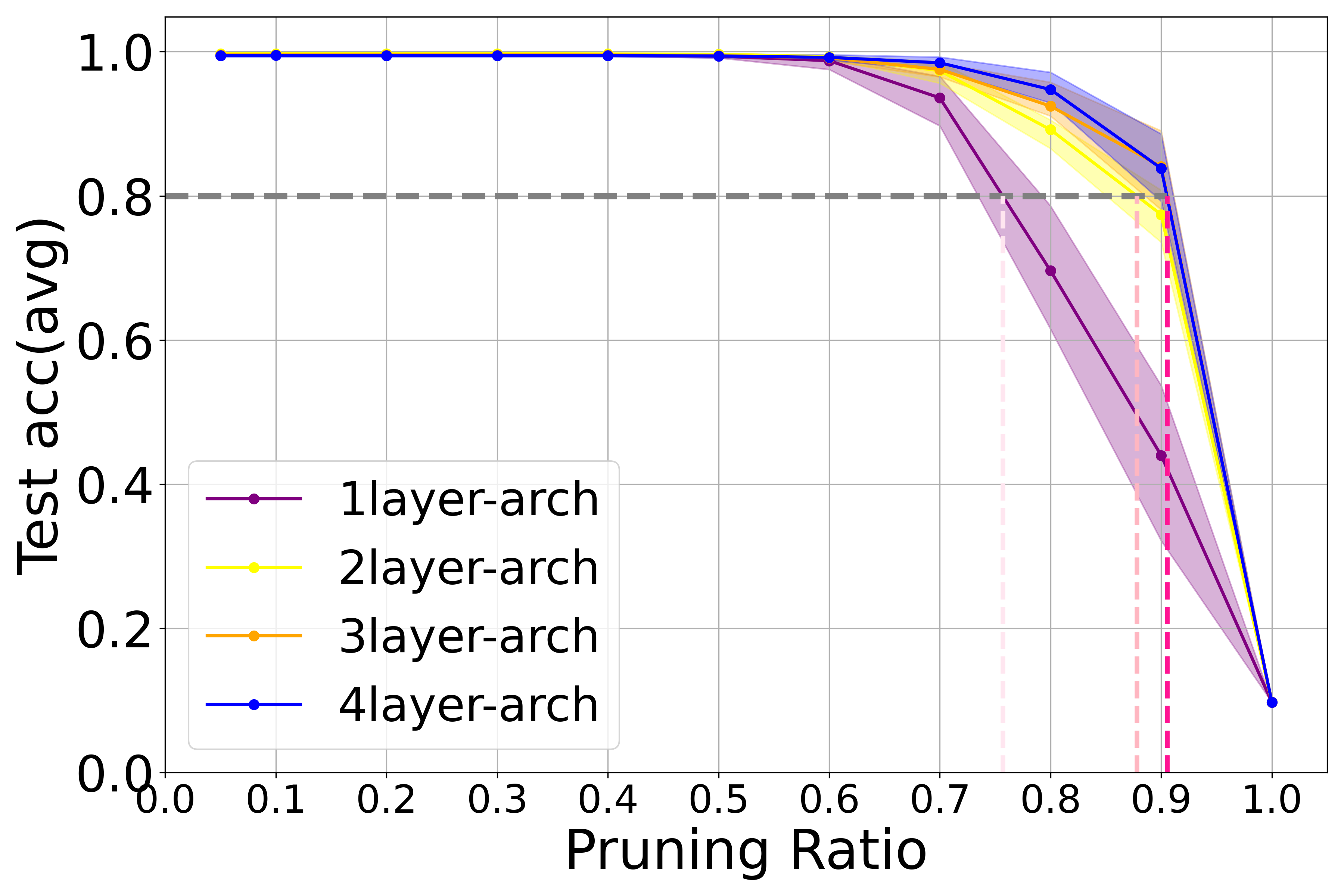}}~\quad
            & \raisebox{-0.5\height}{\includegraphics[width=0.28\textwidth] {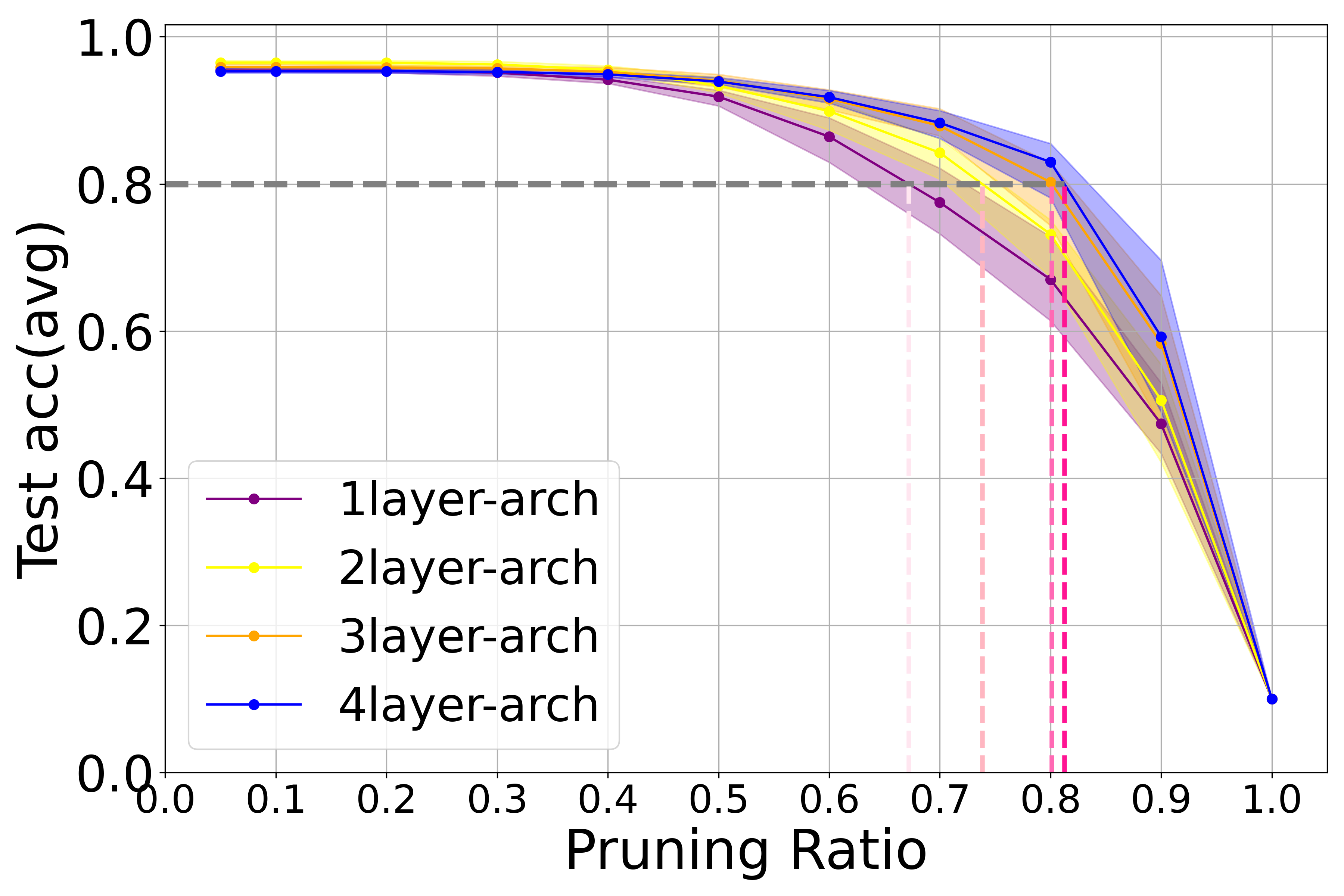}}~\quad
            & \raisebox{-0.5\height}{\includegraphics[width=0.28\textwidth]{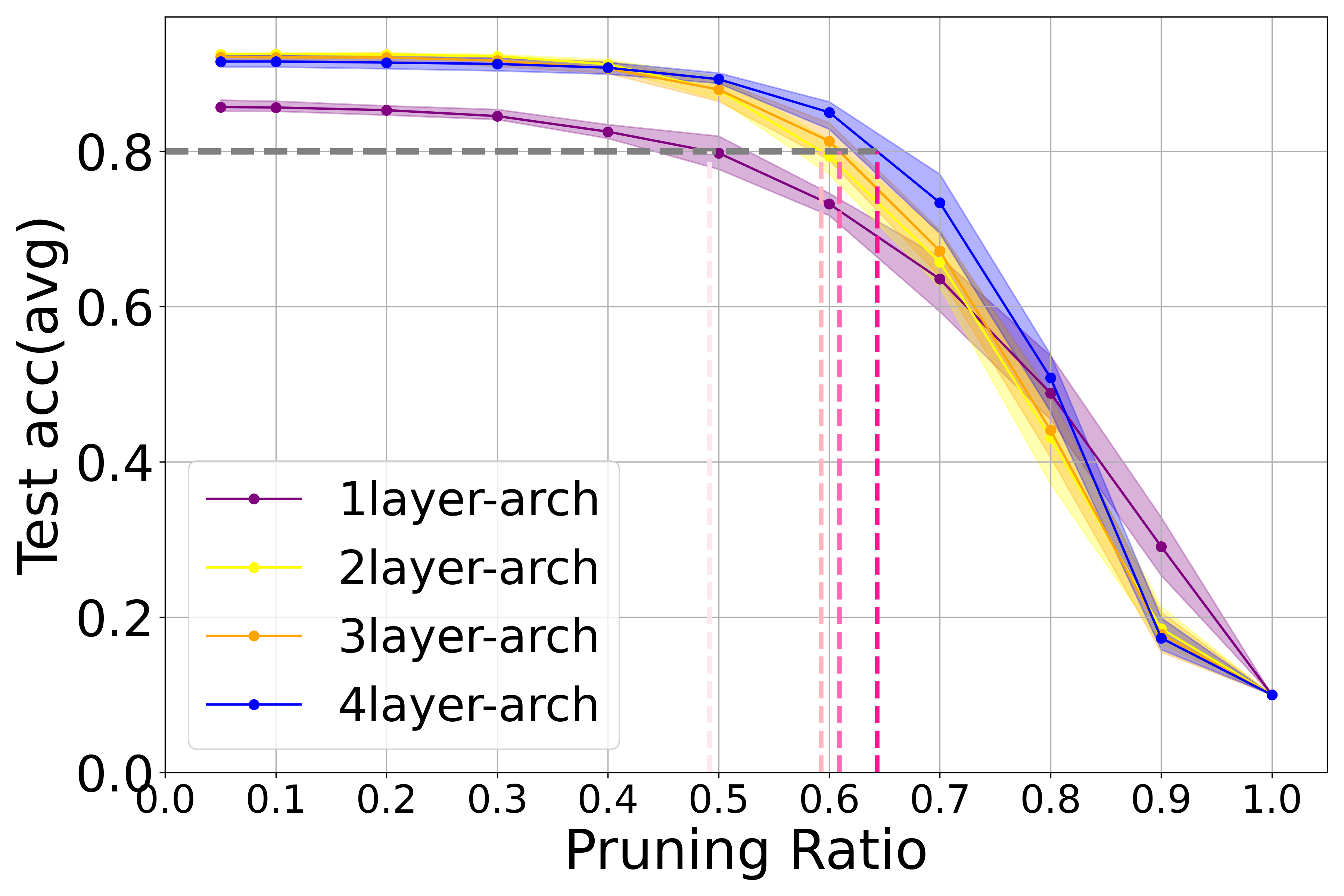}}\\
            %& \raisebox{-0.5\height}{\includegraphics[width=0.24\textwidth]{dnn/pruning_mlp_cifar100_alllayers.png}} \\

        \rotatebox[origin=c]{90}{{\sc Quantization}} 
            & \raisebox{-0.5\height}{\includegraphics[width=0.28\textwidth]{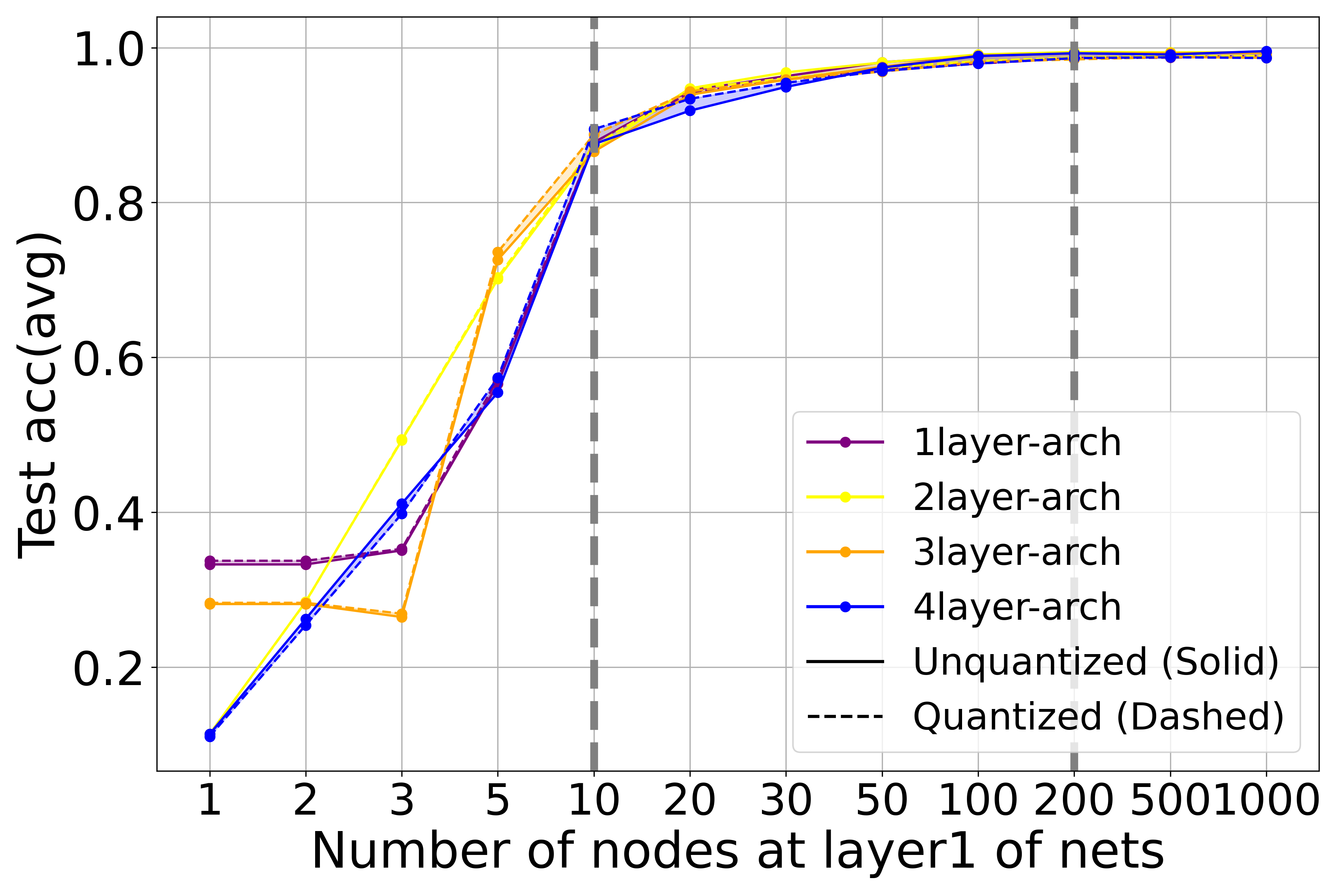}}~\quad
            & \raisebox{-0.5\height}{\includegraphics[width=0.28\textwidth] {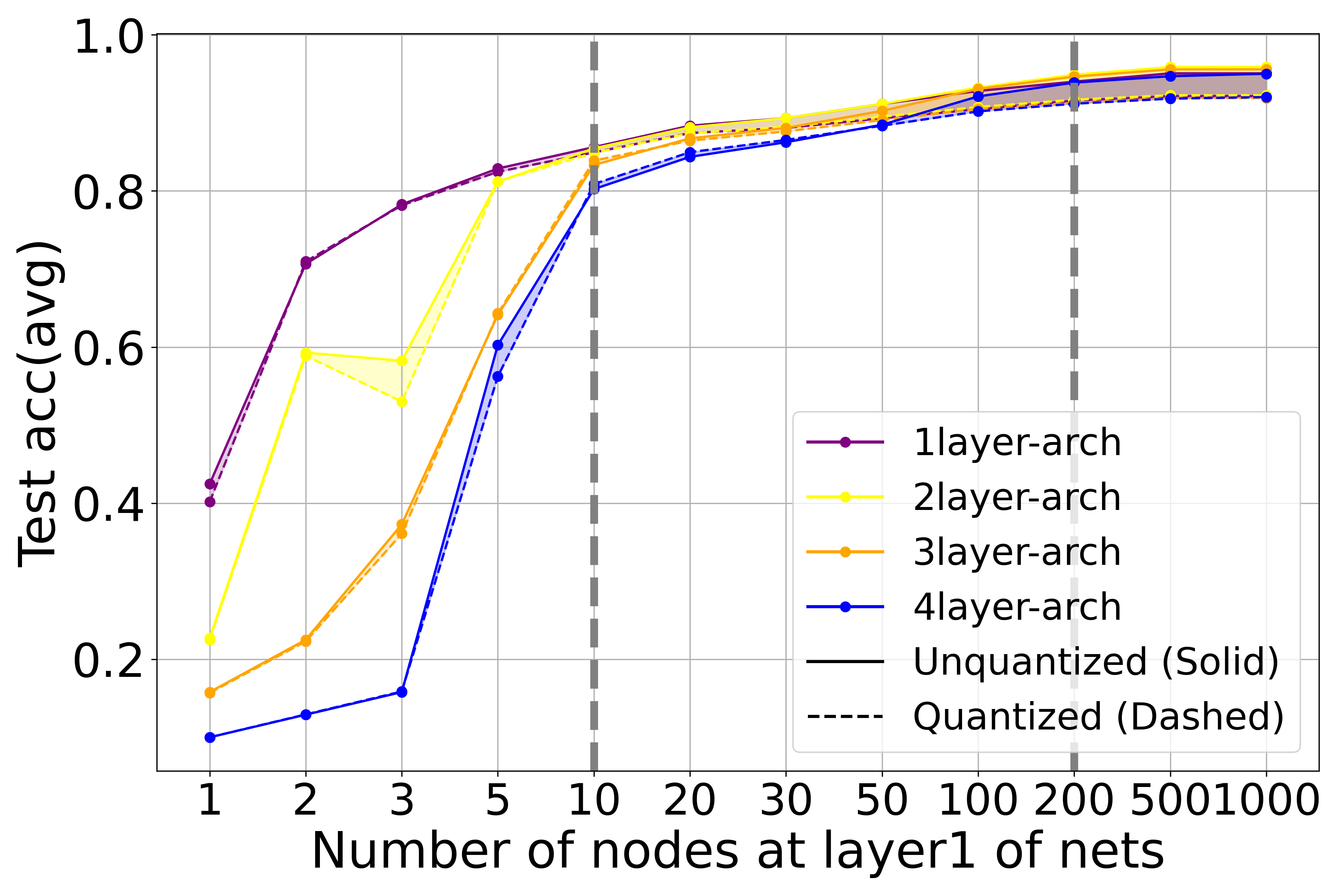}}~\quad
            & \raisebox{-0.5\height}{\includegraphics[width=0.28\textwidth]{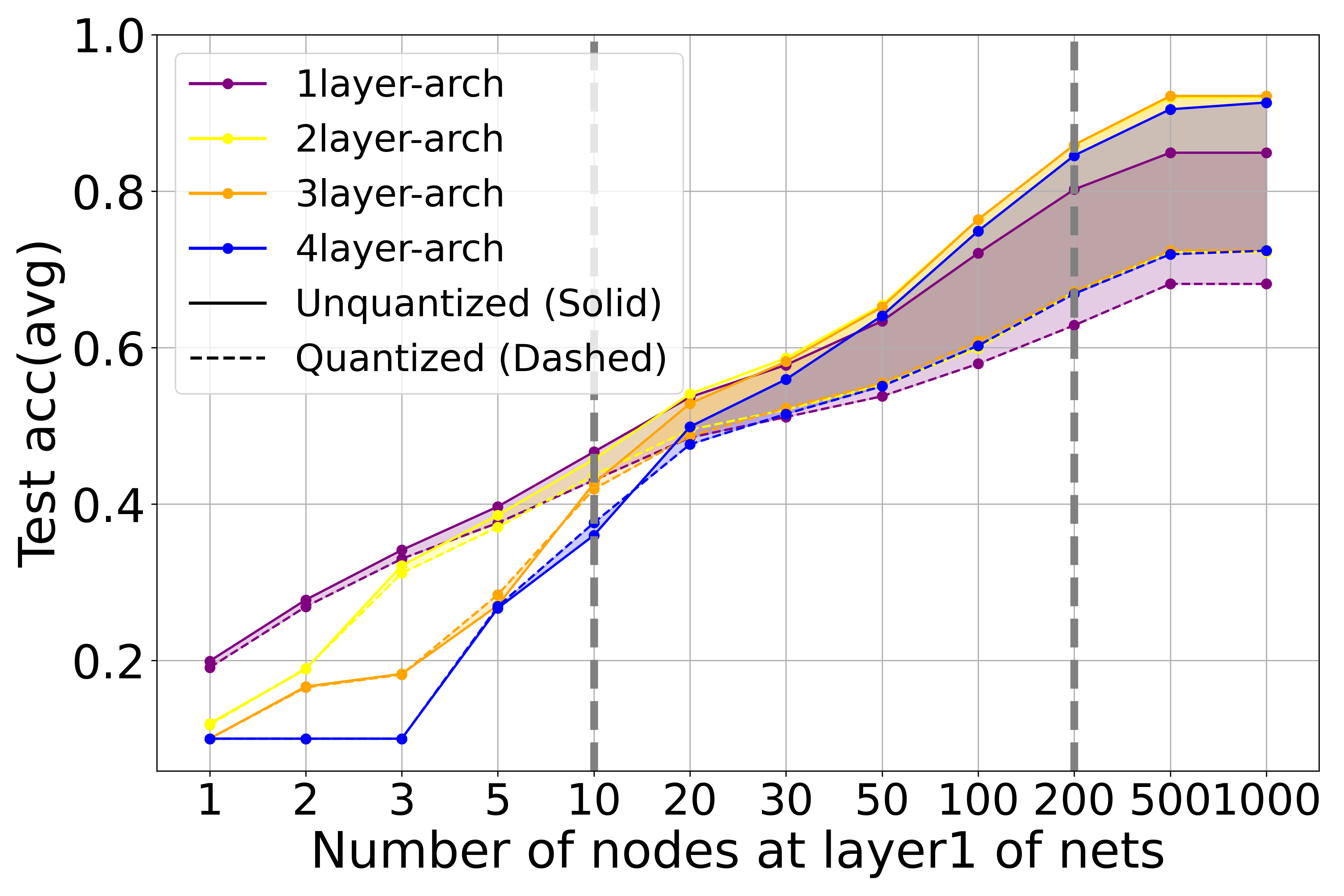}}\\
            %& \raisebox{-0.5\height}{\includegraphics[width=0.24\textwidth]{dnn/quantization_cifar100_alllayers.png}}
    \end{tabular}
    \caption{Convergence, Pruning, and Quantization Results on DNN. The solid lines indicate the average test accuracy, and the shaded area is $\pm$ standard deviation. Colors represent the network with hidden layers of $1,$ $ $2,$ $ $3,$ and $4,$ respectively. The curves reflect changes in network performance and stability as the number of nodes in the first hidden layer increases ($ x$-axis in Convergence and Quantization). A value 5 on the $x-axis$ corresponds to architectures \{$5$\}, \{$5\times5$\}, \{$5\times5\times5$\}, and \{$5\times5\times5\times5$\} respectively to 1, 2, 3, and 4, layered DNNs, and the input and output layers varied respectively to datasets. For pruning, the $x-axis$ represents the pruning ratio, and dotted vertical lines indicate the pruning ratio that matches the curves for different layers when the same accuracy is achieved. Symbol $\theta$ indicates average learnable parameters.}
    \label{fig:results_dnn}
\end{figure*}

\begin{table}[t]
\caption{Summary of minimum average stable parameter counts, safe pruning ratio, and 8-bit quantization gap.}% for different architectures and datasets.}
\label{tab:stability_pruning_quant}
\centering
\setlength{\tabcolsep}{3pt}   % 列间距稍微紧一点
\scriptsize                  % 比 \small 更适合单栏窄表
\begin{tabular}{ccccc}
\toprule
Dataset & Architecture & Min. Params (Stability) & Safe Pruning \% & 8-bit Gap \% \\
\midrule
\multirow{3}{*}{MNIST}
  & DNN & $2\times10^{4}$ & 60 & 0.15 \\
  & CNN & $3\times10^{4}$ & 20 & 0.74 \\
  & ViT & $5\times10^{3}$ & 20 & 10.66 \\
\midrule
\multirow{3}{*}{F-MNIST}
  & DNN & $2\times10^{4}$ & 40 & 1.42 \\
  & CNN & $3\times10^{4}$ & 20 & 1.10 \\
  & ViT & $5\times10^{3}$ & 20 & 3.22 \\
\midrule
\multirow{3}{*}{CIFAR-10}
  & DNN & $2\times10^{6}$ & 40 & 7.50 \\
  & CNN & $3\times10^{6}$ & 5 & 14.80 \\
  & ViT & $6\times10^{3}$ & 10 & 4.17 \\
\bottomrule
\end{tabular}
\end{table}

\paragraph{Convergence profile for stable learning}
Across MLPs/DNNs, CNNs, and ViTs on image datasets, we observe a common two-stage learning pattern. Small-capacity models lie in an unstable region, with low mean accuracy and high variance. Once the network's width is sufficient, the models enter a stable region where performance variance decreases and accuracy stabilizes. On MNIST, exhaustive experiments with MLPs/DNNs indicate that above 1,000 learnable parameters, variance performance decreases, and accuracy stabilizes. CNNs reach stability with a modest number of feature channels on MNIST and Fashion MNIST, typically when the first layer provides about four to eight channels, at which point test accuracy is about $99.4\%$ to $99.8\%$ on MNIST and about $99.0\%$ to $99.6\%$ on Fashion MNIST. On CIFAR-10, the same trend holds, but the stable region shifts to larger channel budgets, with accuracy continuing to rise from sixteen to thirty-two channels and reaching about 83\% to 88\%. ViTs require a larger embedding width to attain similar stability. On MNIST, a near-plateau appears only at the largest width, with accuracy around $96\%.$ On Fashion MNIST, the rise is slower and saturates later, around $0.80$ at the largest width. On CIFAR-10, accuracy increases steadily across the explored range and reaches about $0.37$ at the largest width, indicating that the models remain under-parameterized within this sweep. These findings show a progression in which MLPs/DNNs stabilize first on simple images, CNNs stabilize with modest channel counts and perform better at mid-level complexity, and ViTs require the largest width before accuracy becomes predictable as a function of the number of learnable parameters.

\begin{figure*}[h!]
    %\footnotesize
    \setlength{\tabcolsep}{10pt}
    \centering
    \begin{tabular}{ccccc}
    \centering
         & {\sc MNIST} & {\sc Fashion MNIST} & {\sc CIFAR-10} \\
        \rotatebox[origin=c]{90}{{\sc Convergence}}
            & \raisebox{-0.5\height}{\includegraphics[width=0.25\textwidth]{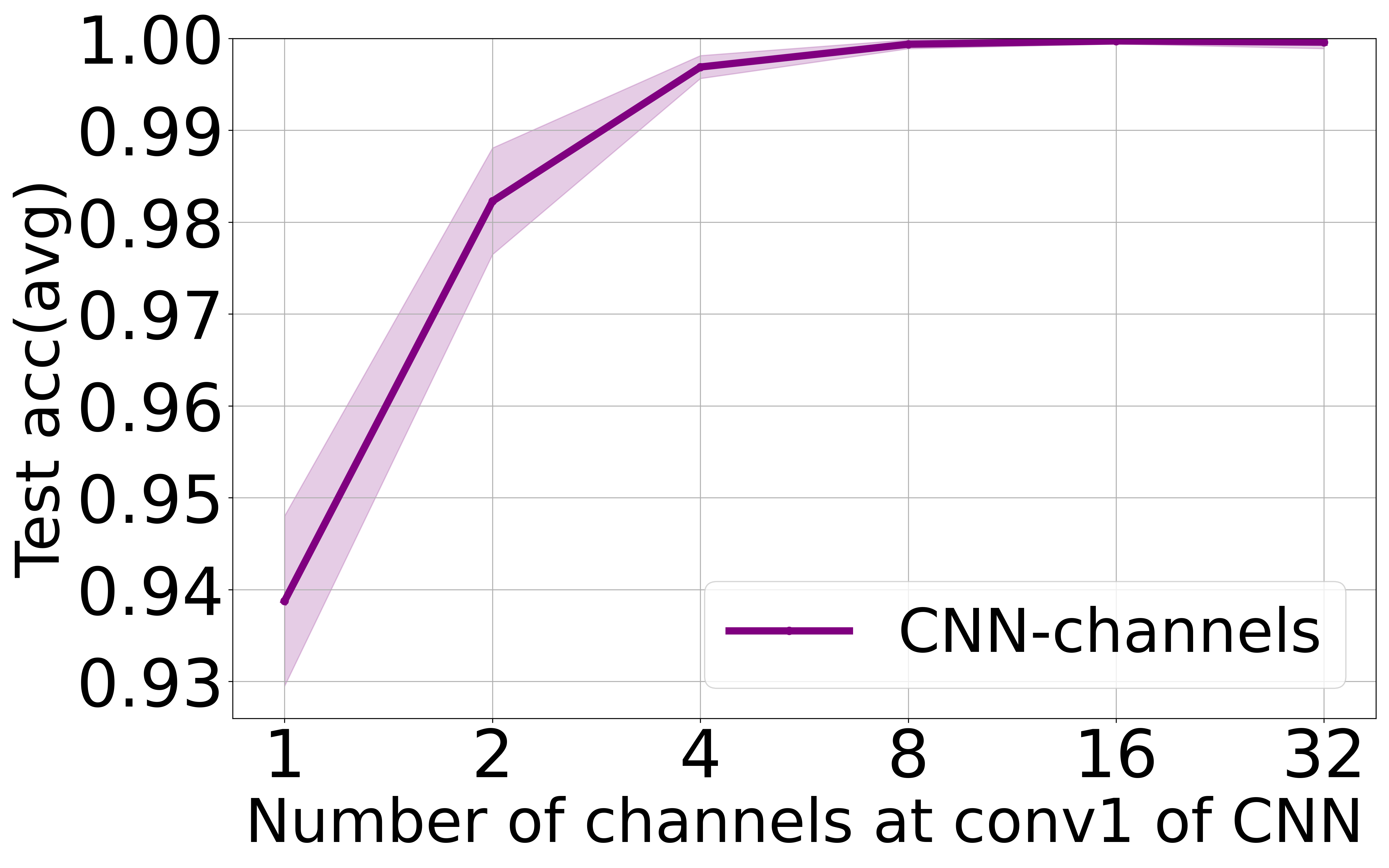}}~\qquad
            & \raisebox{-0.5\height}{\includegraphics[width=0.25\textwidth] {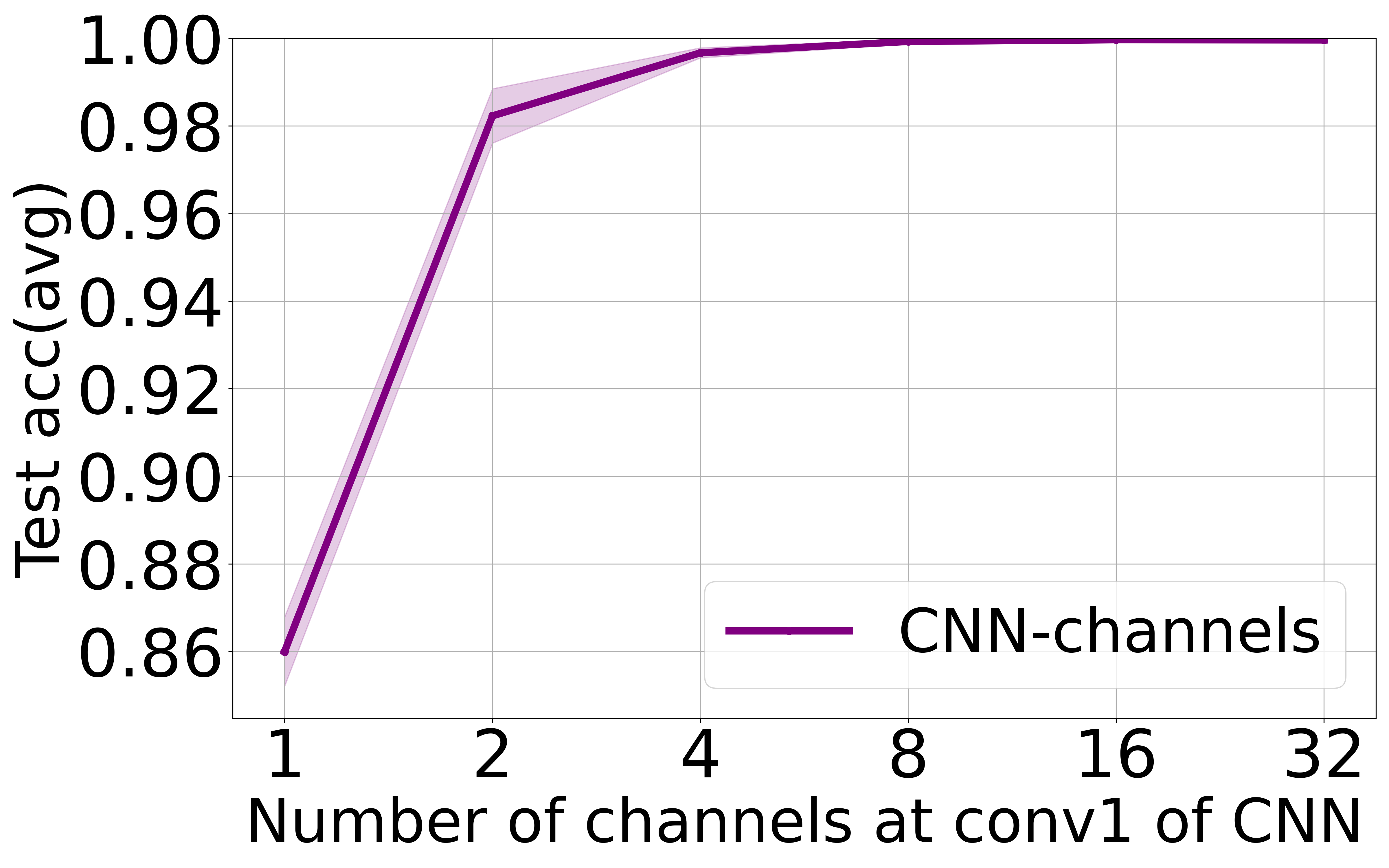}}~\qquad
            & \raisebox{-0.5\height}{\includegraphics[width=0.25\textwidth]{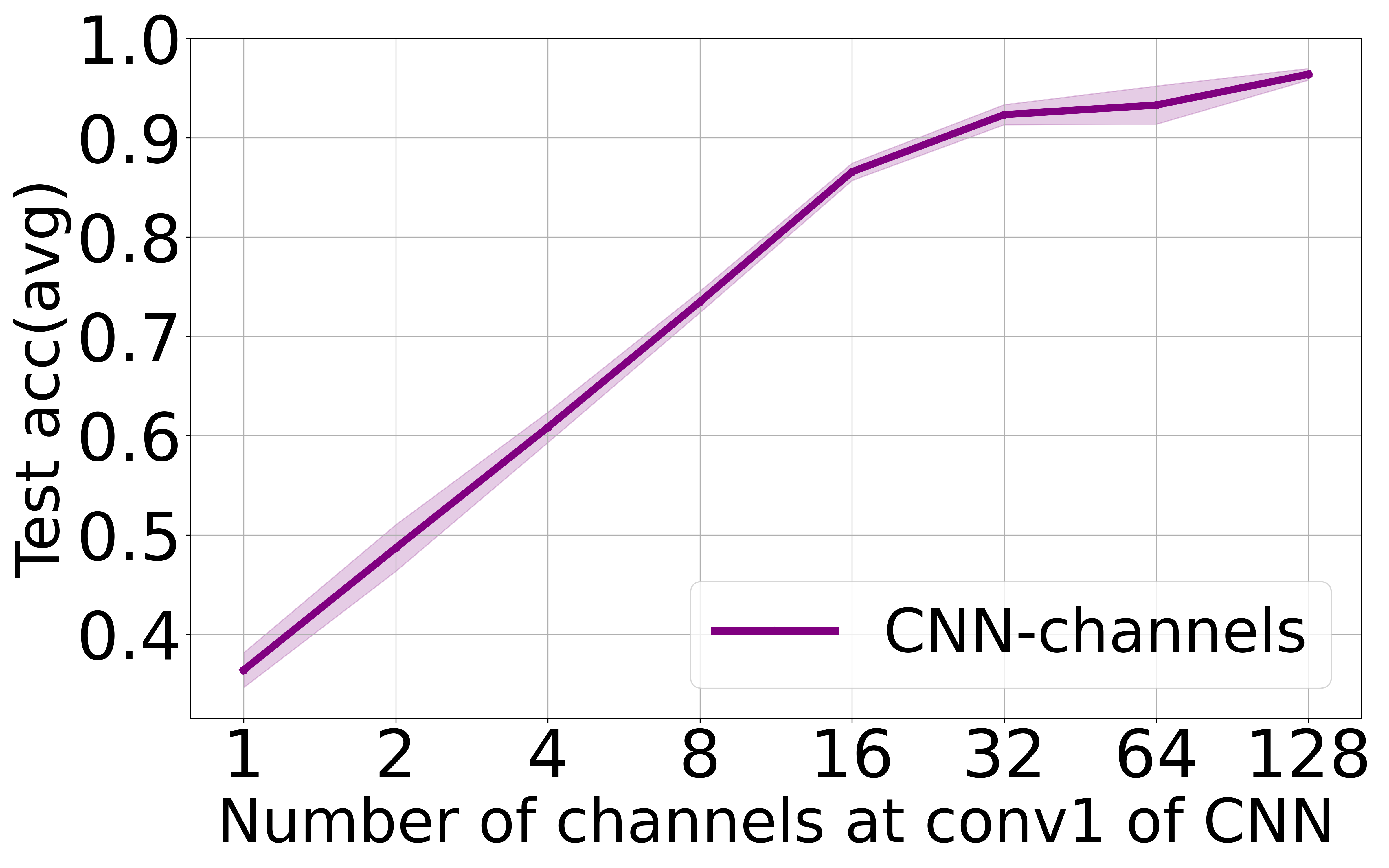}}\\
        
        \rotatebox[origin=c]{90}{{\sc Pruning}} 
            & \raisebox{-0.5\height}{\includegraphics[width=0.25\textwidth]{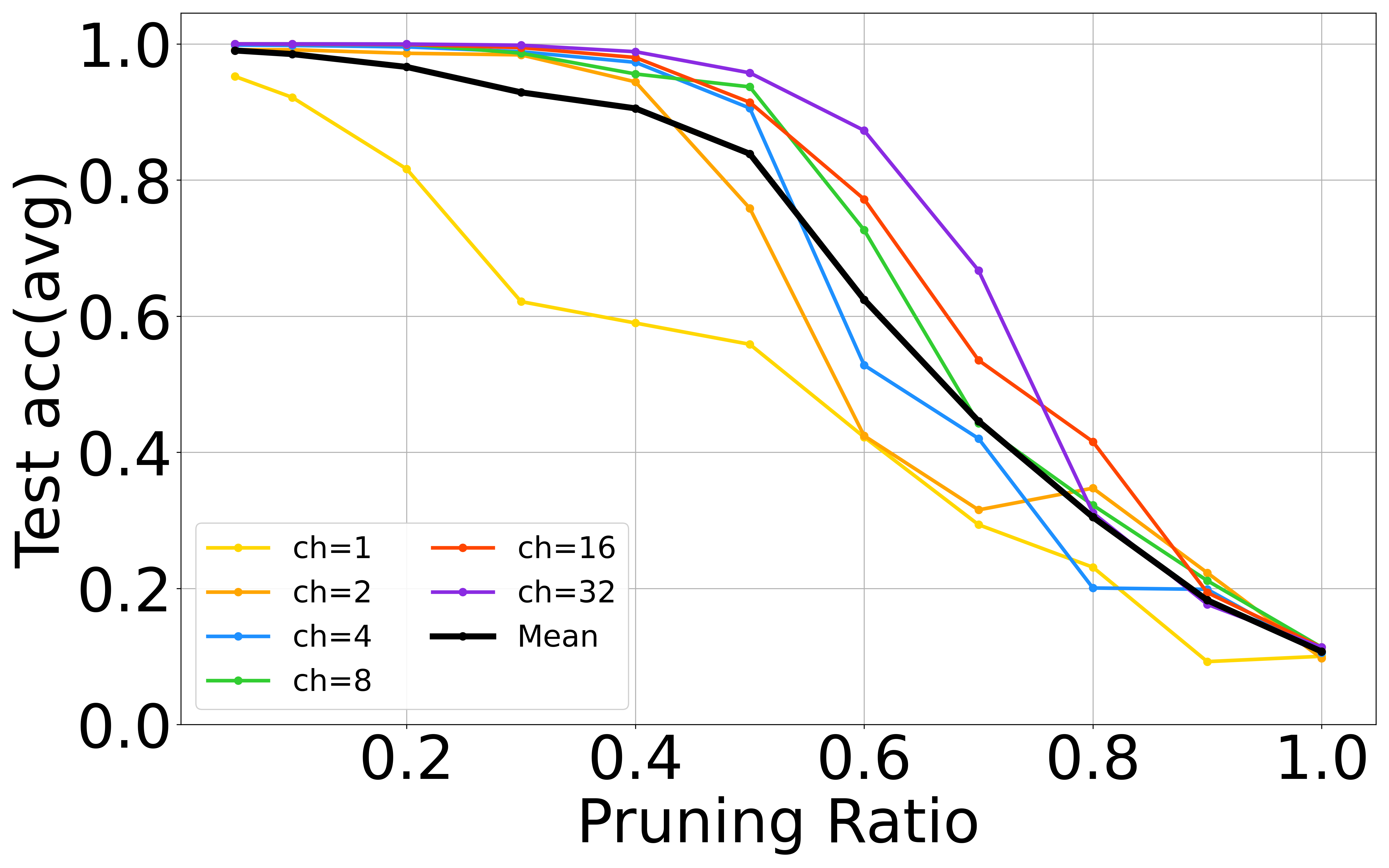}}~\qquad
            & \raisebox{-0.5\height}{\includegraphics[width=0.25\textwidth] {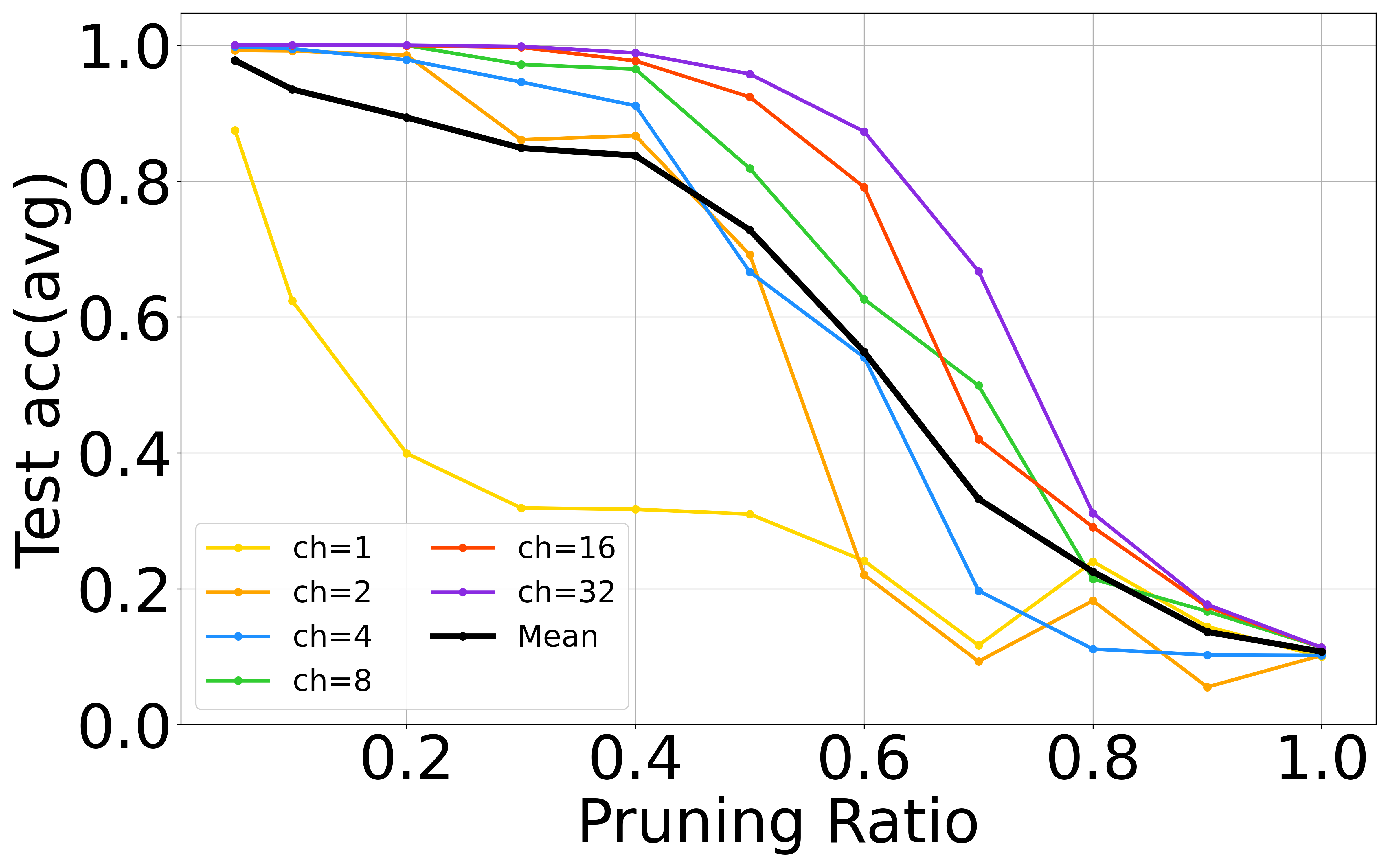}}~\qquad
            & \raisebox{-0.5\height}{\includegraphics[width=0.25\textwidth]{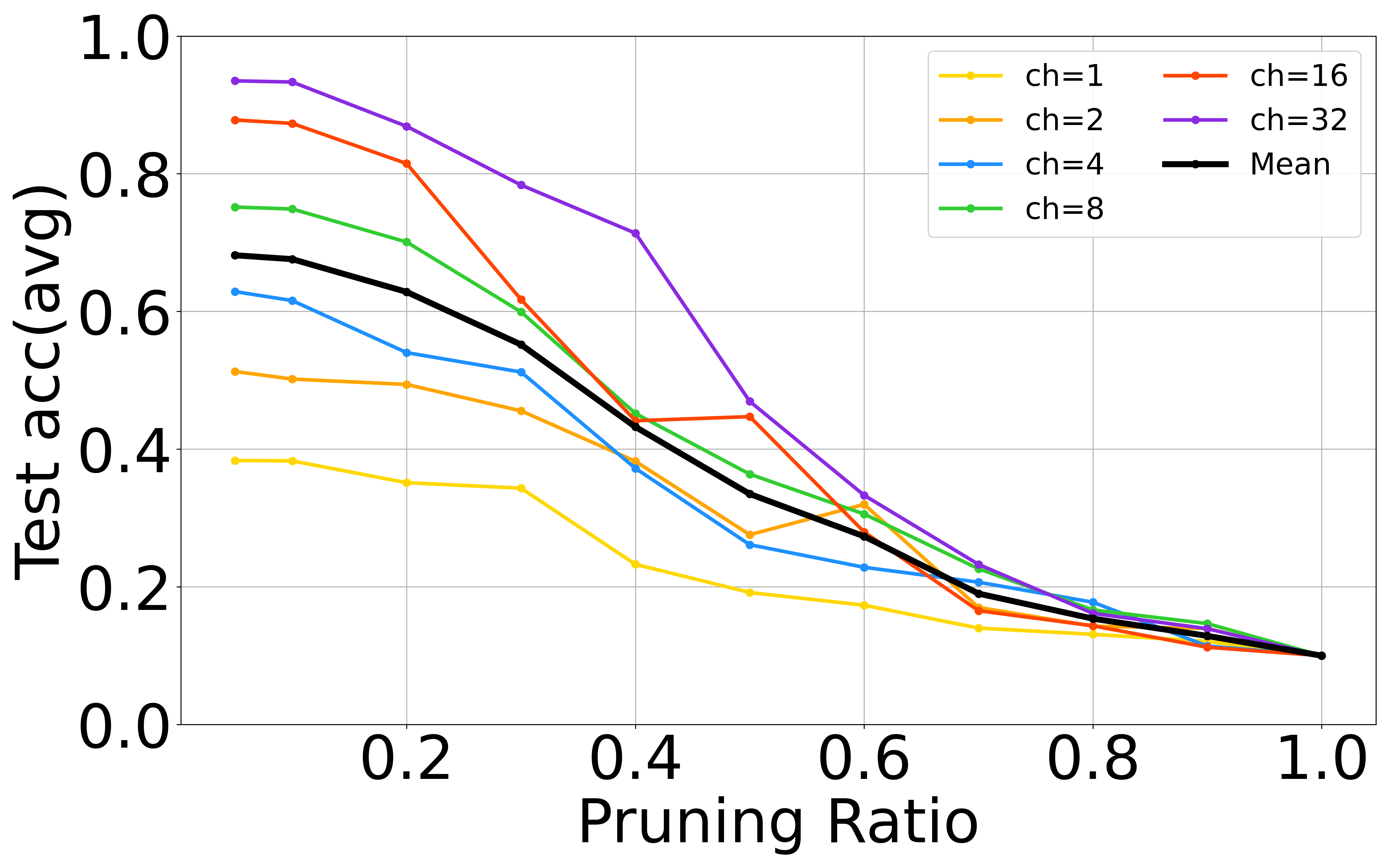}}\\

        \rotatebox[origin=c]{90}{{\sc Quantization}} 
            & \raisebox{-0.5\height}{\includegraphics[width=0.25\textwidth]{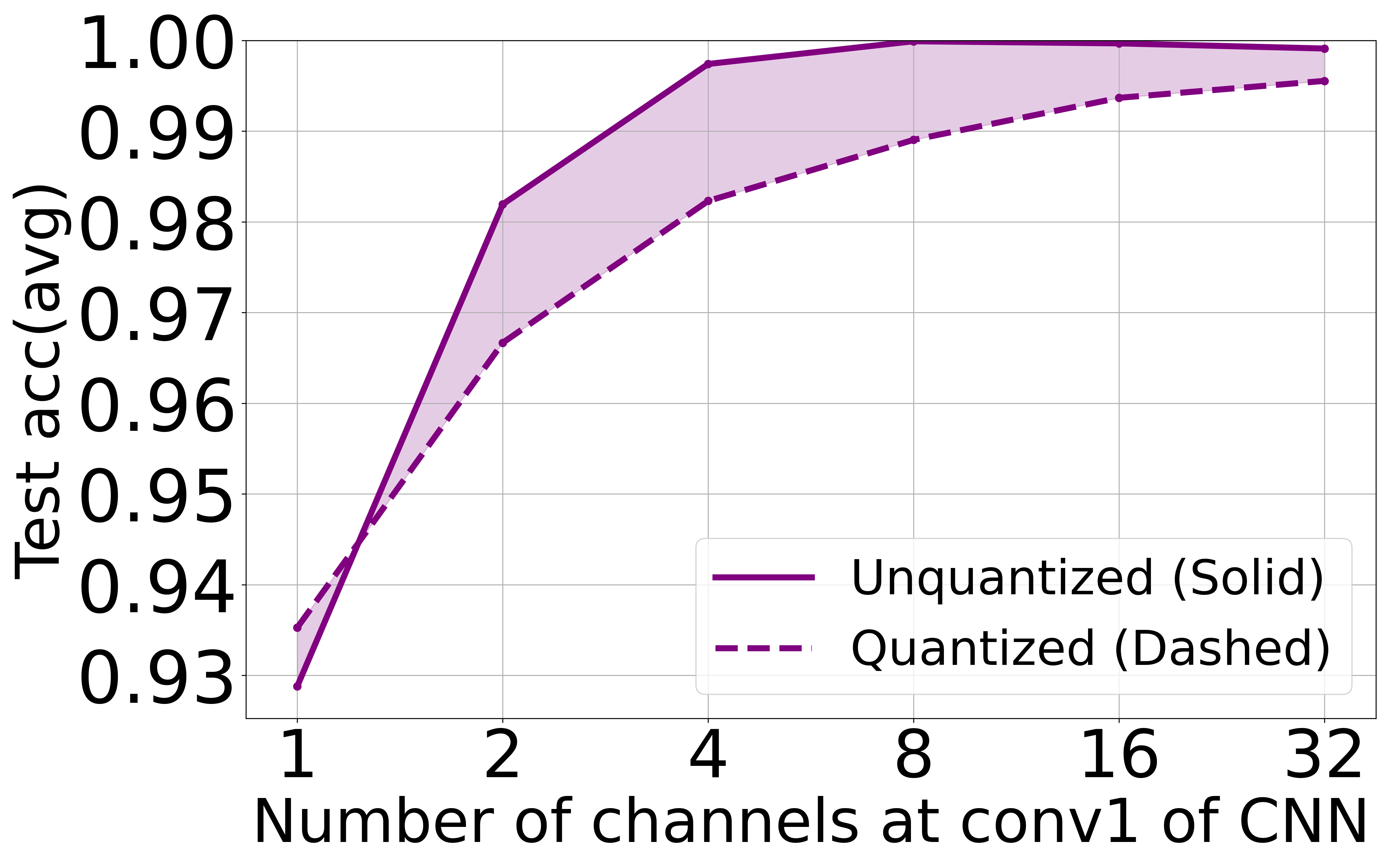}}~\qquad
            & \raisebox{-0.5\height}{\includegraphics[width=0.25\textwidth] {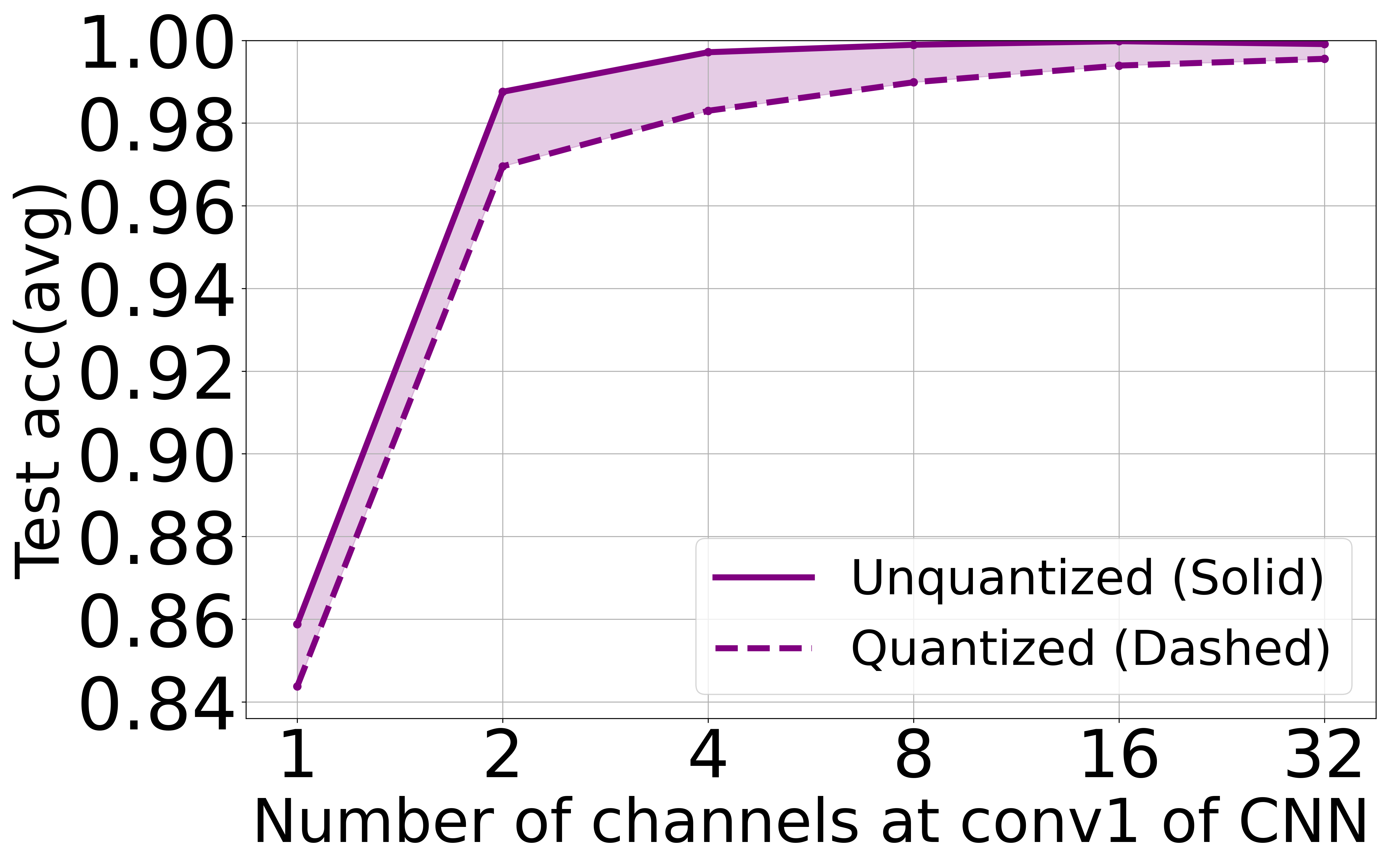}}~\qquad
            & \raisebox{-0.5\height}{\includegraphics[width=0.25\textwidth]{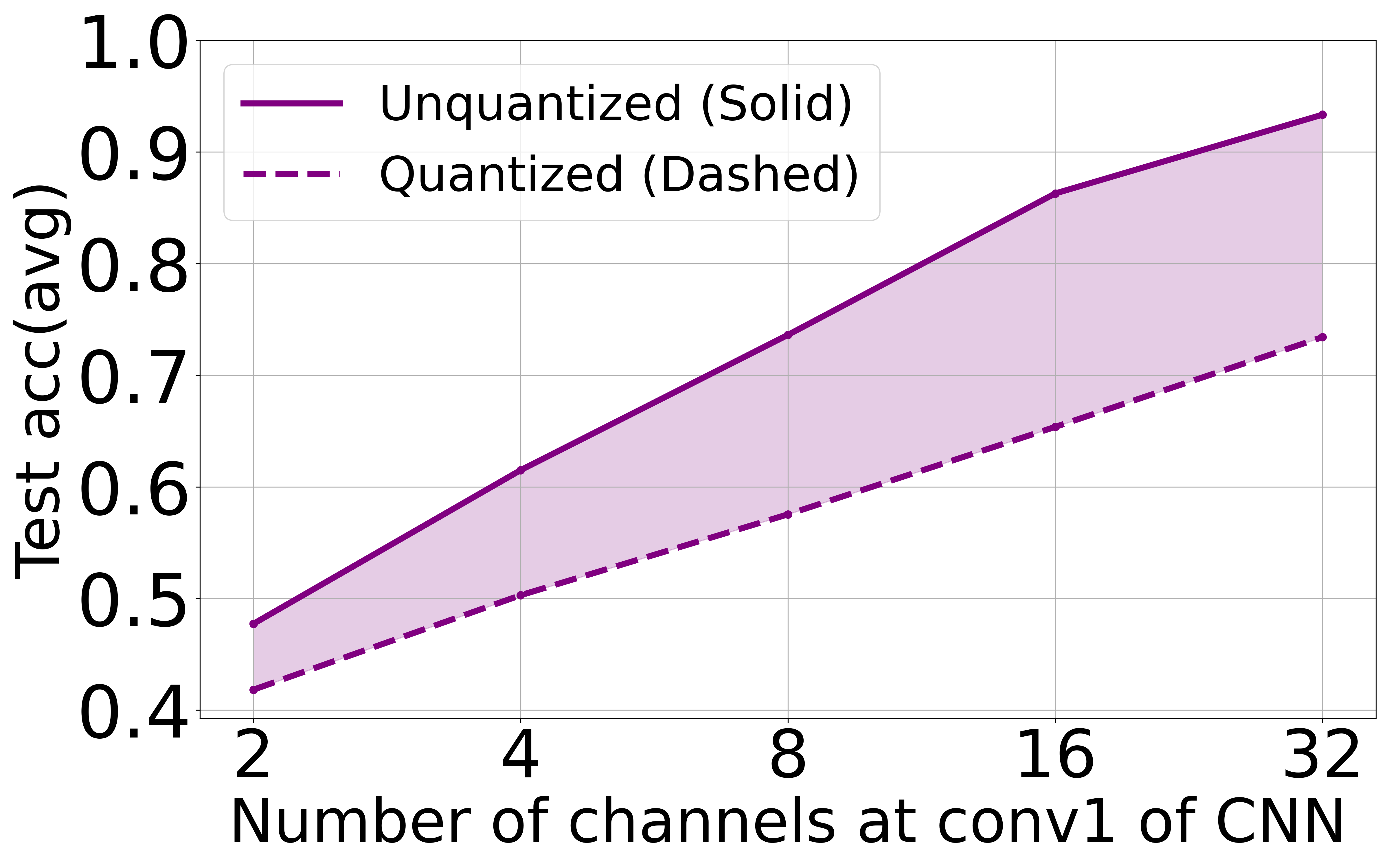}}
    \end{tabular}
    \caption{Convergence, Pruning, and Quantization Results on CNN. The curves reflect network performance and stability changes as the number of channels in the first convolution layer increases ($x-axis$ in Convergence and Quantization).}
    \label{fig:results_cnn}
\end{figure*}

\begin{figure*}[h!]
    %\footnotesize
    \setlength{\tabcolsep}{10pt}
    \centering
    \begin{tabular}{cccc}
    \centering
         & {\sc MNIST} & {\sc Fashion MNIST} & {\sc CIFAR-10} \\
        \rotatebox[origin=c]{90}{{\sc Convergence}}
            & \raisebox{-0.5\height}{\includegraphics[width=0.25\textwidth]{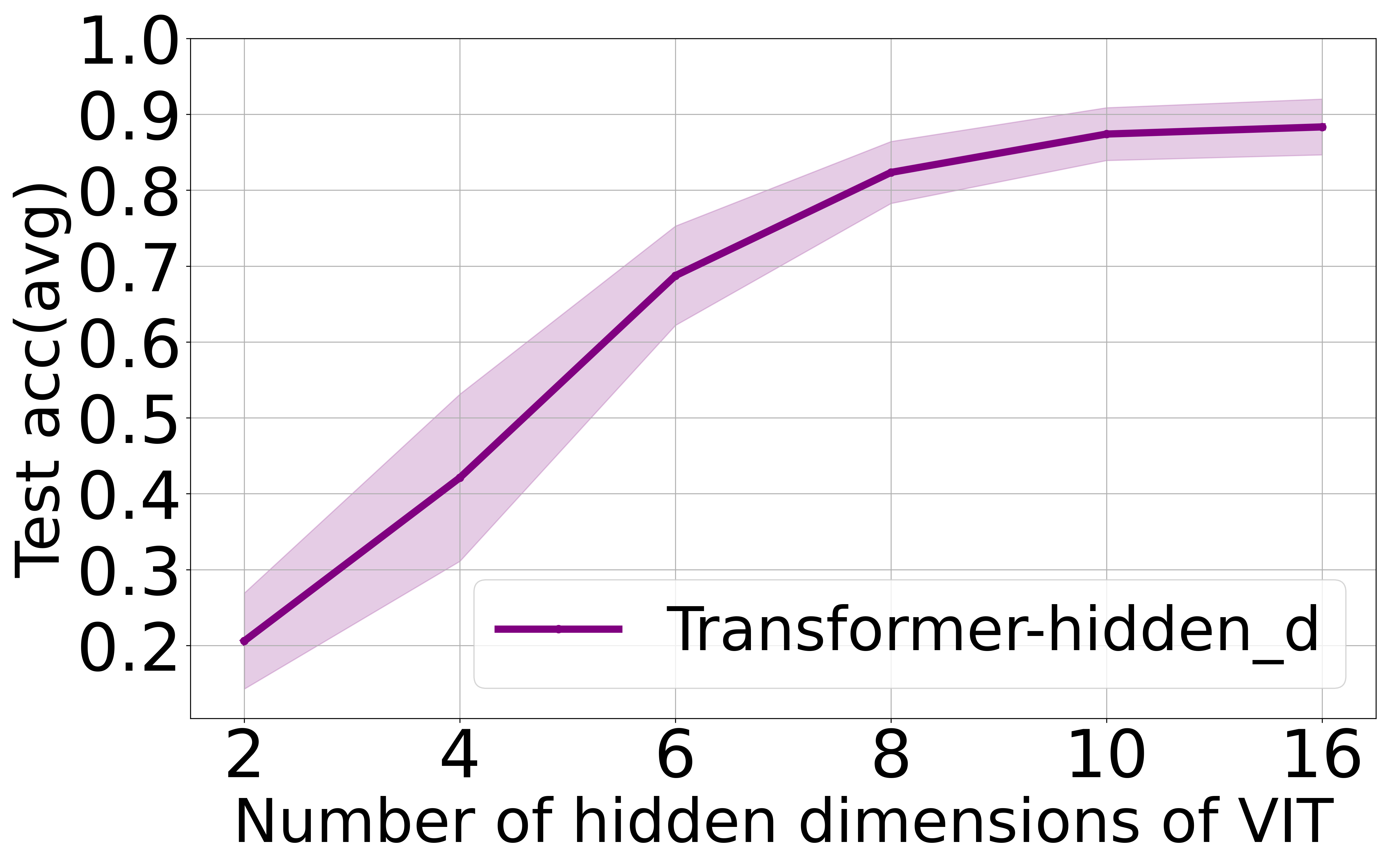}}~\qquad
            & \raisebox{-0.5\height}{\includegraphics[width=0.25\textwidth] {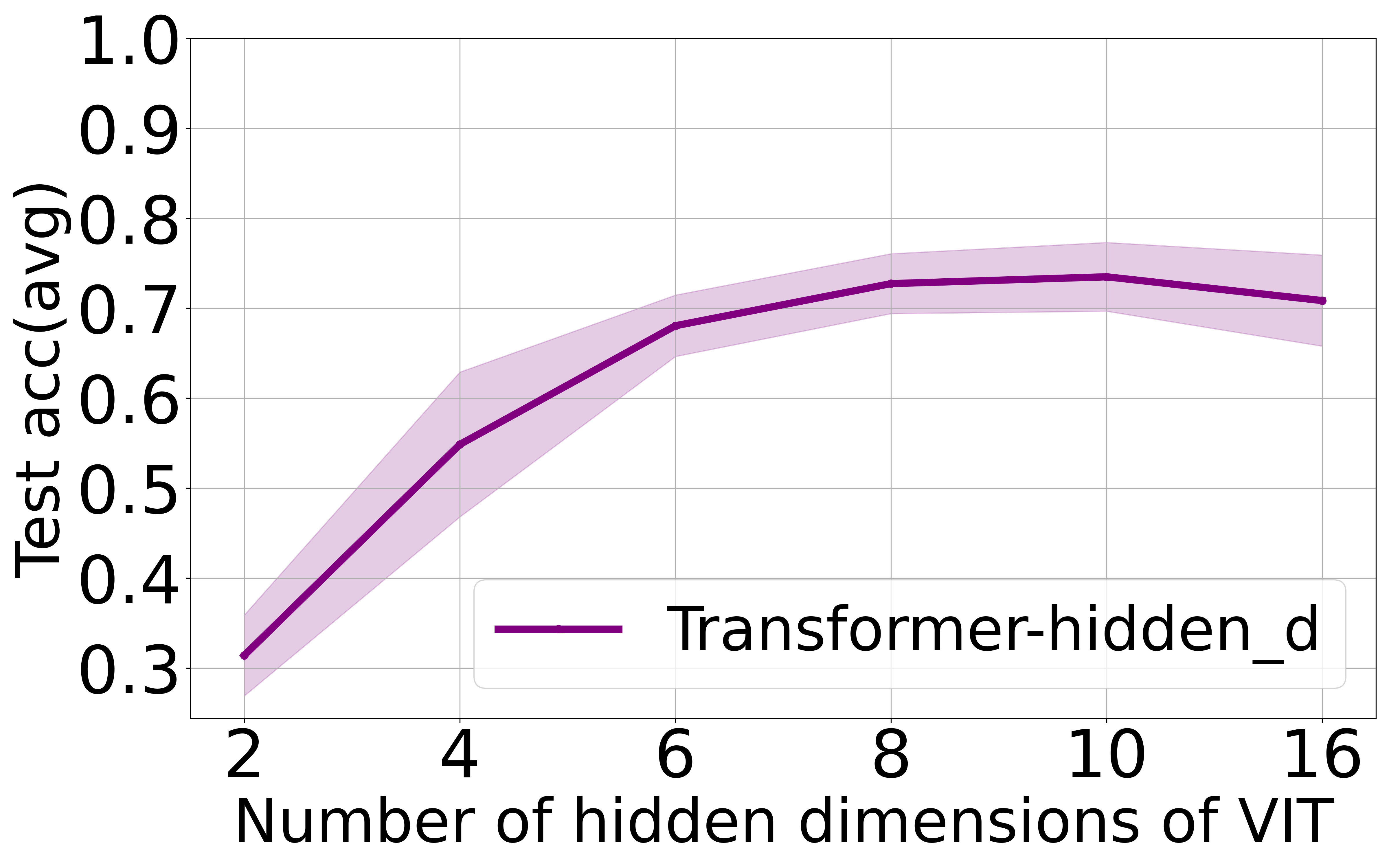}}~\qquad
            & \raisebox{-0.5\height}{\includegraphics[width=0.25\textwidth]{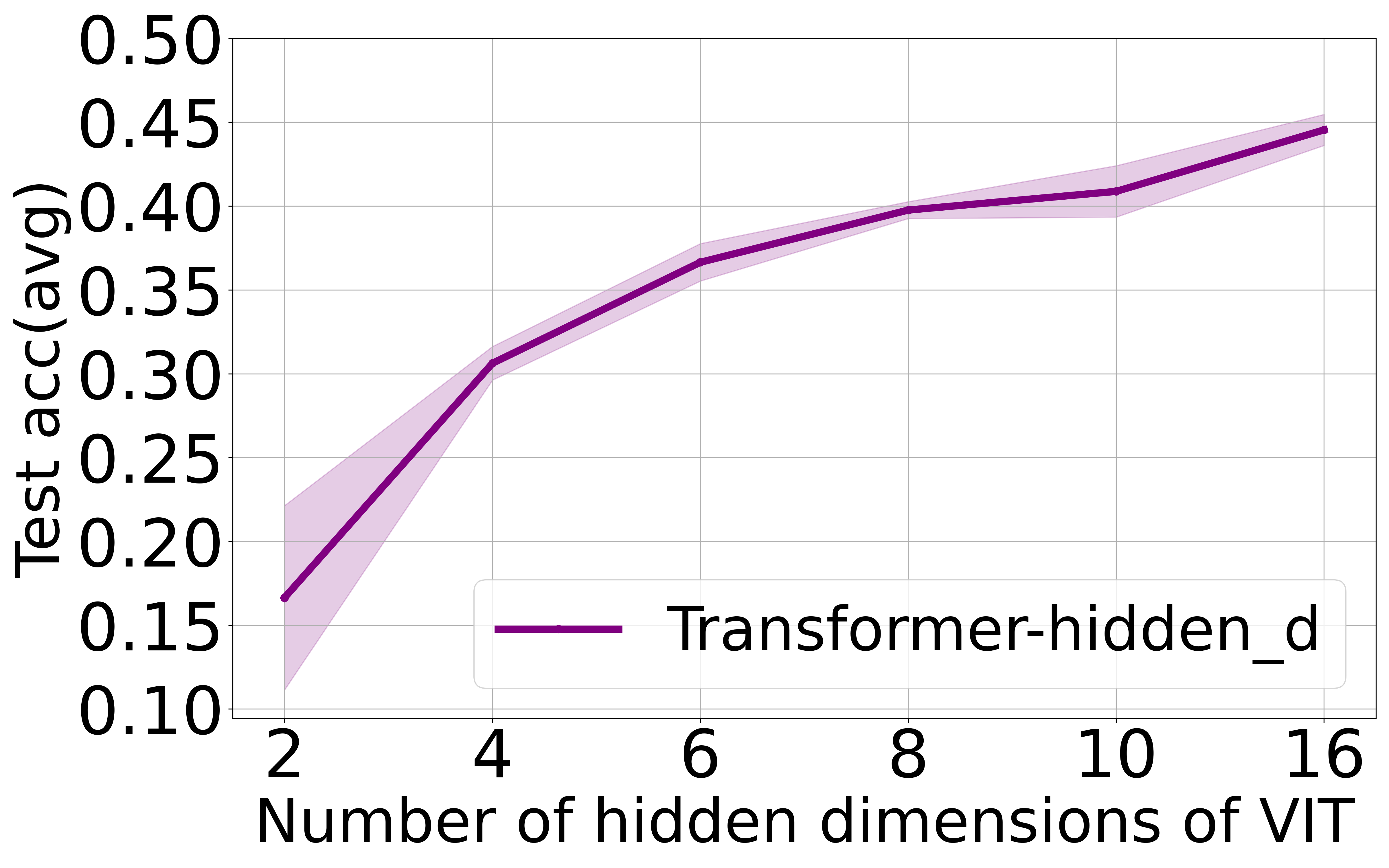}}\\
        
        \rotatebox[origin=c]{90}{{\sc Pruning}} 
            & \raisebox{-0.5\height}{\includegraphics[width=0.25\textwidth]{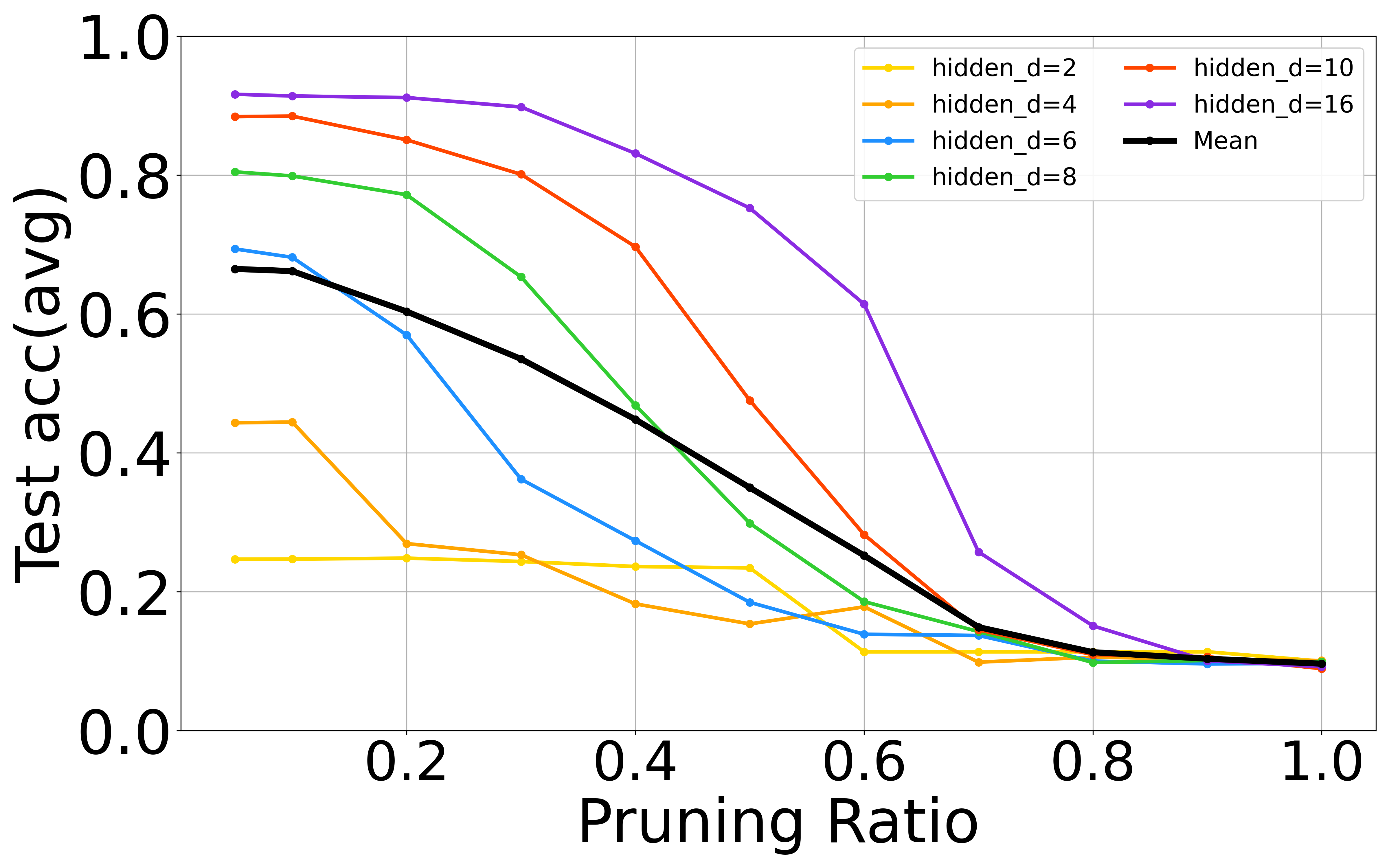}}~\qquad
            & \raisebox{-0.5\height}{\includegraphics[width=0.25\textwidth] {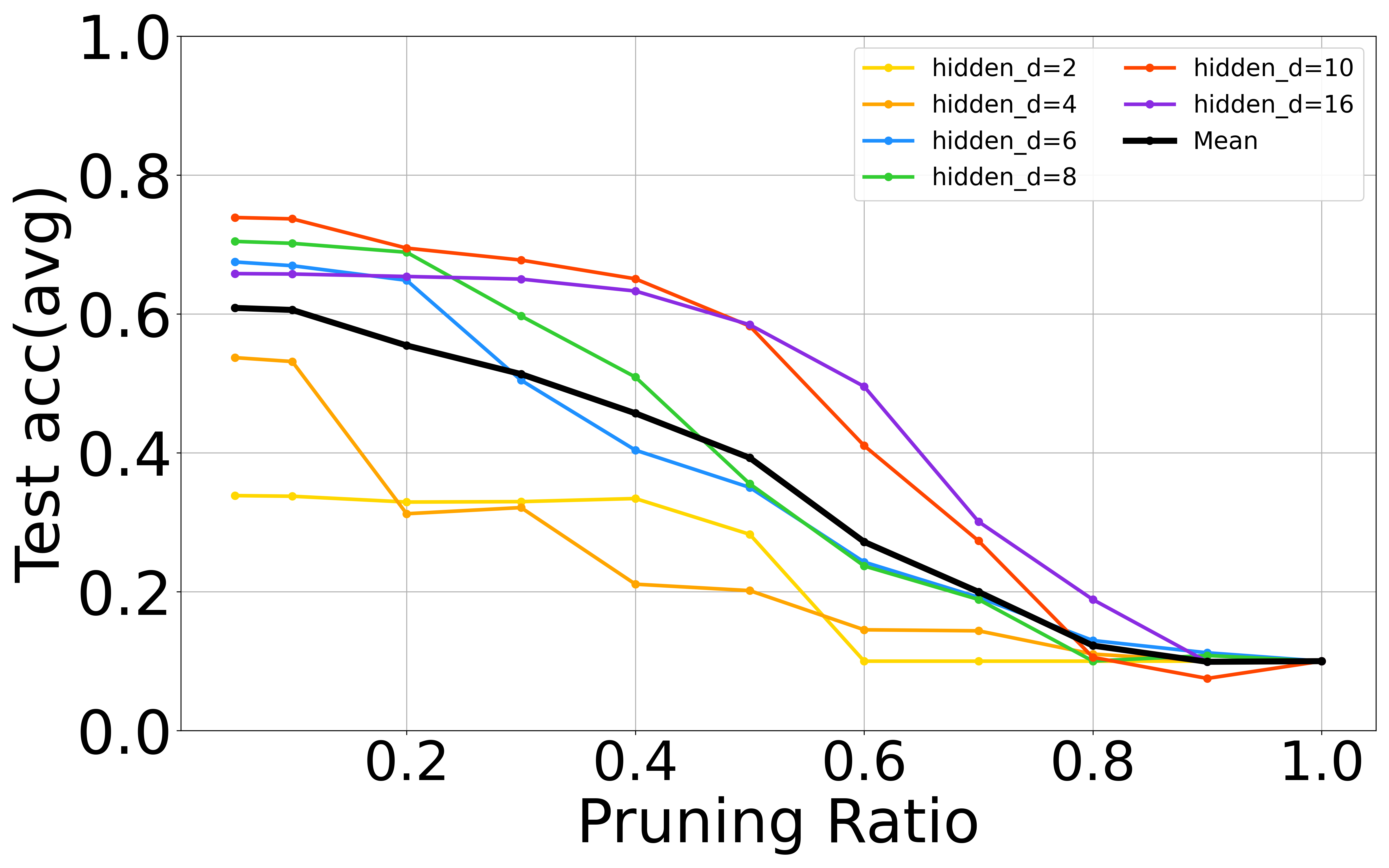}}~\qquad
            & \raisebox{-0.5\height}{\includegraphics[width=0.25\textwidth]{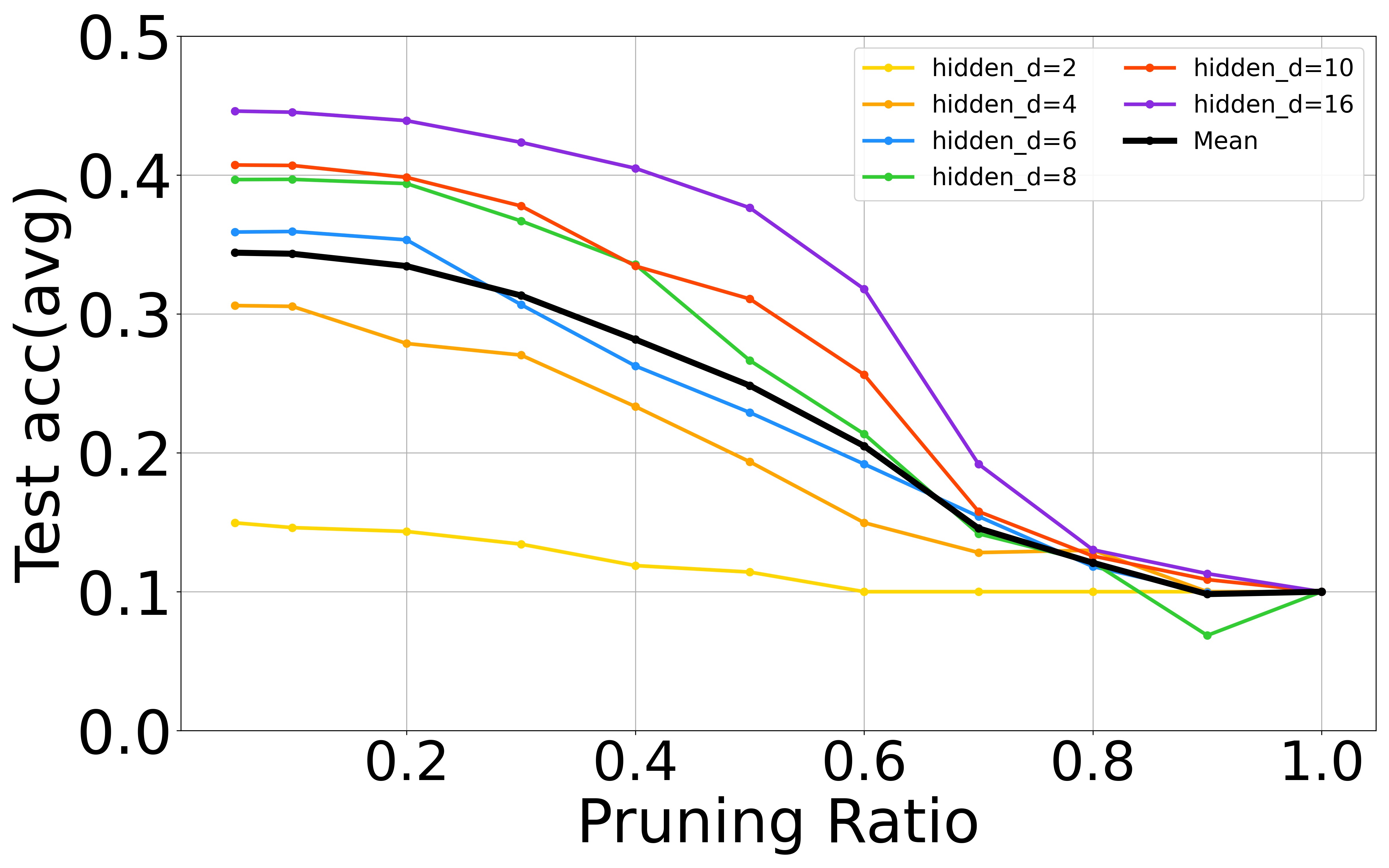}}\\

        \rotatebox[origin=c]{90}{{\sc Quantization}} 
            & \raisebox{-0.5\height}{\includegraphics[width=0.25\textwidth]{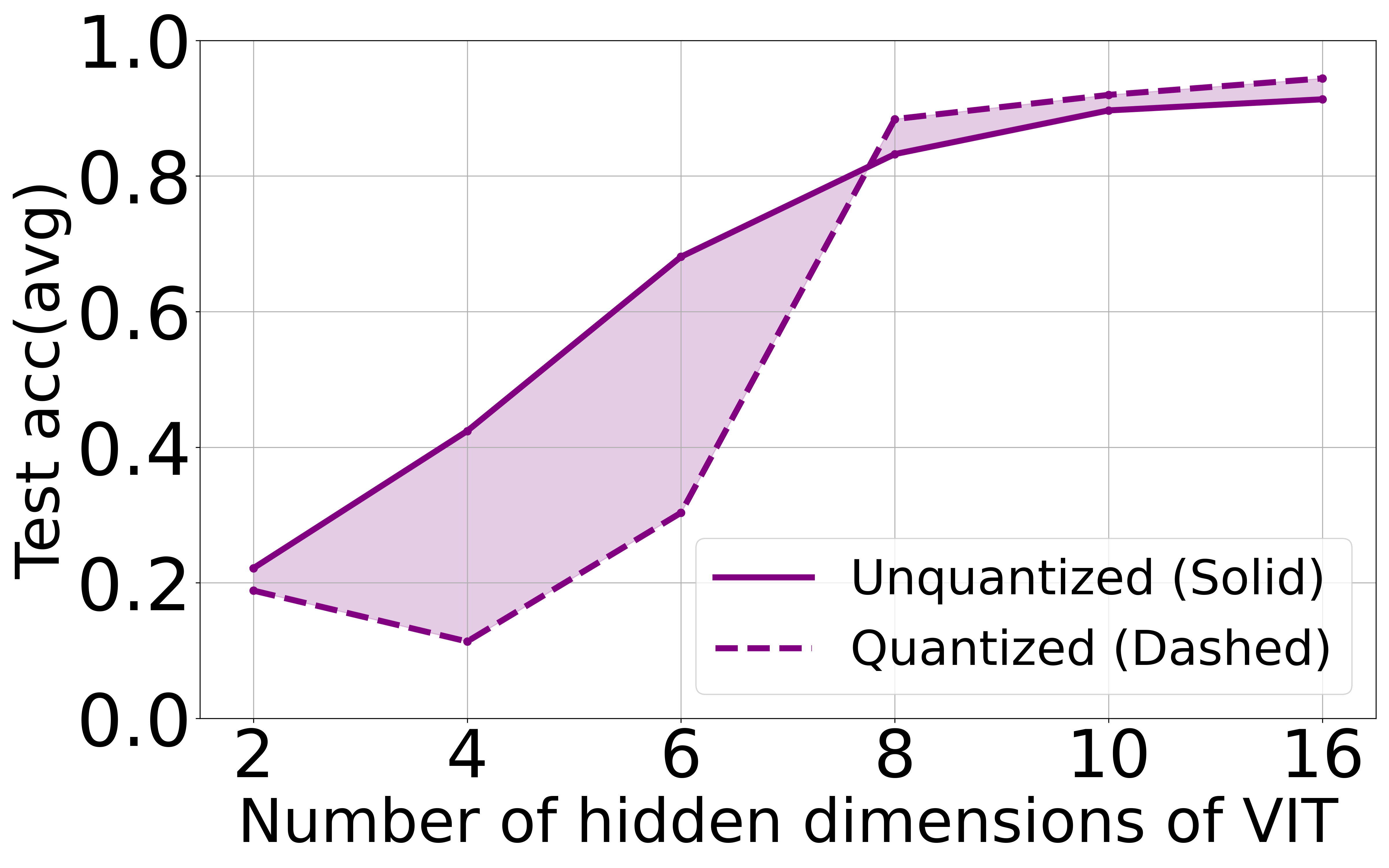}}~\qquad
            & \raisebox{-0.5\height}{\includegraphics[width=0.25\textwidth] {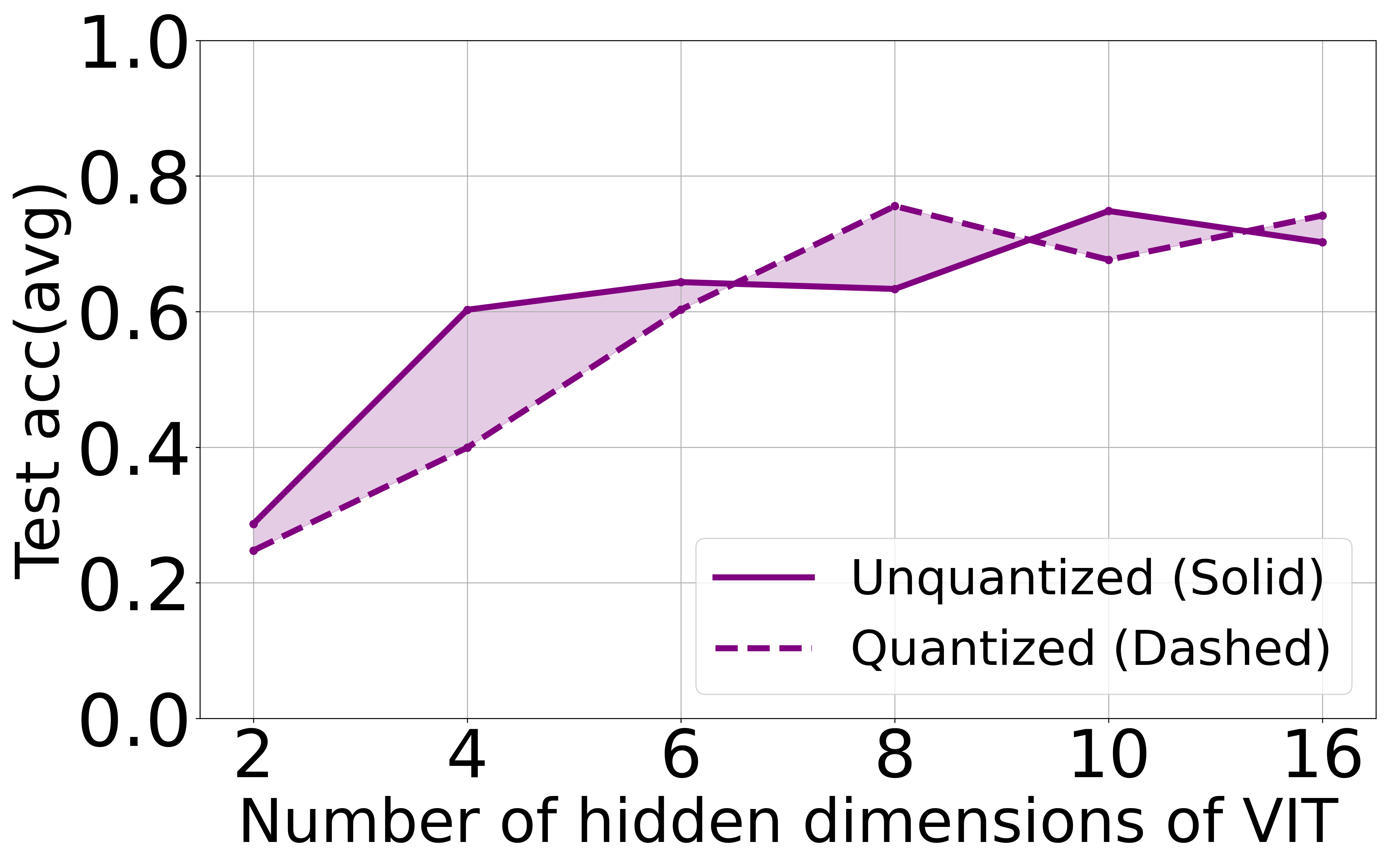}}~\qquad
            & \raisebox{-0.5\height}{\includegraphics[width=0.25\textwidth]{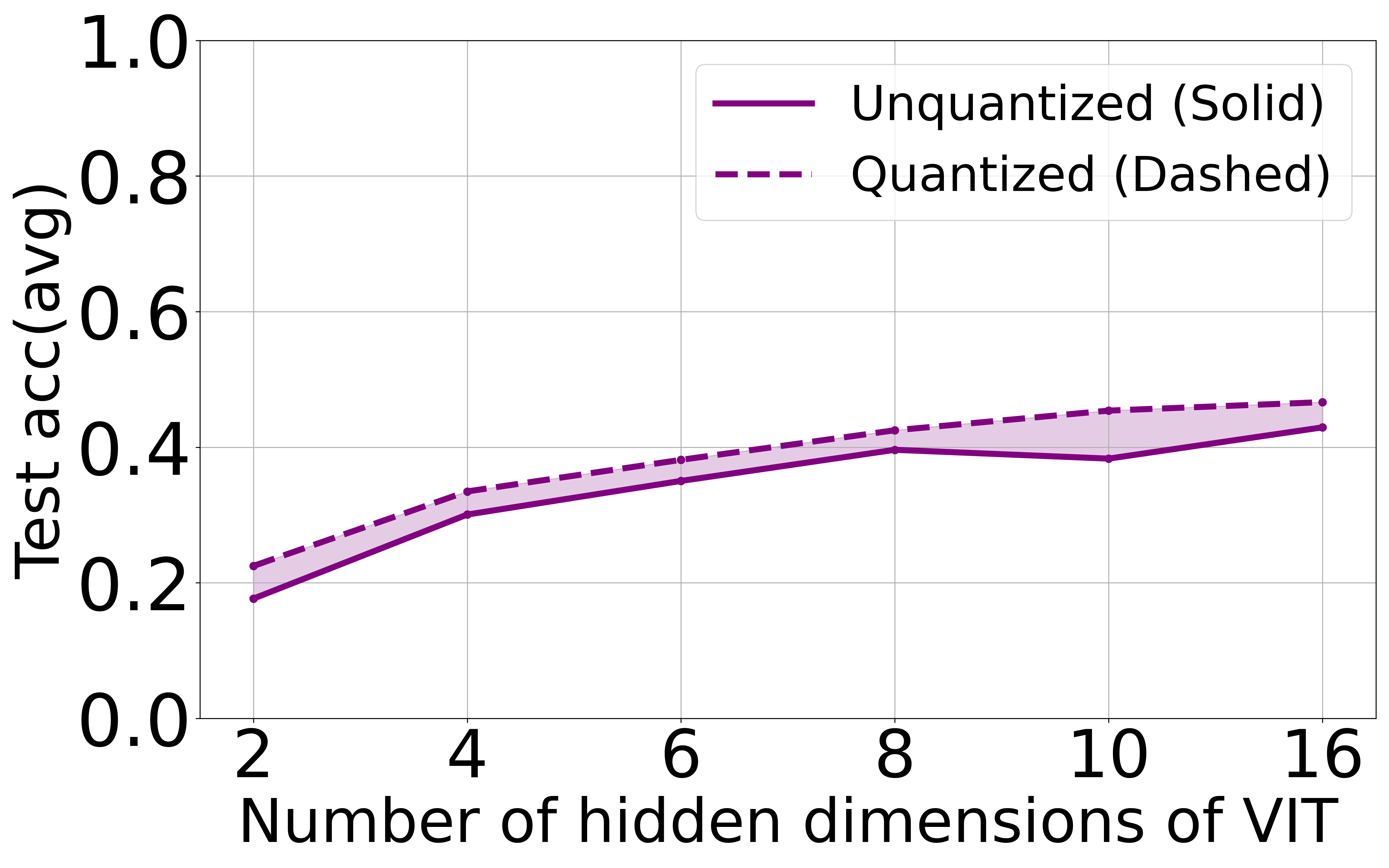}}
    \end{tabular}
    \caption{Convergence, Pruning, and Quantization Results on VIT. The curves reflect network performance and stability changes as the number of embedding channels increases ($x-axis$ in Convergence and Quantization).}
    \label{fig:results_vit}
    \vspace{-2em}
\end{figure*}

\paragraph{Pruning profile for stable learning}
Pruning reveals three regimes shared across MLPs/DNNs, CNNs, and ViTs. At low pruning ratios, accuracy remains close to the unpruned level, indicating substantial learnable parameter redundancy. In the intermediate regime, accuracy declines sharply and performance variance increases. In the high-pruning regime, accuracy collapses because the remaining parameters cannot sustain the task. On MNIST, MLPs/DNNs remain near baseline up to $50\%-60\%$ pruning, and then show a clear decline beyond about $60\%-70\%$ pruning. CNNs on MNIST and Fashion MNIST stay close to the baseline up to moderate pruning and begin to fail between $60\%$ and $70\%,$ whereas on CIFAR-10, the decline occurs earlier and often lies between $50\%$ and $60\%.$ ViTs follow the same pattern but are more sensitive at a given accuracy level. At a small hidden dimension, the failing region begins at lower pruning ratios, around $45\%-60\%$ on MNIST and $35\%-45\%$ on CIFAR-10. Within each model family, wider or deeper configurations tolerate higher pruning before failure. These results indicate that redundancy becomes effective only after the models have entered the stable learning region and that the safe pruning window grows with architectural capacity.

\paragraph{Quantization profile for stable learning}
Quantization from 32-bit to 8-bit is governed by the margin achieved with full-precision training. When the learned margin is large, 8-bit precision preserves accuracy. When the margin is small, accuracy drops. On MNIST, MLPs/DNNs show little to no performance decline once the parameter count exceeds the stability threshold. For CNNs, the gap between full precision and 8-bit is negligible on MNIST and Fashion MNIST once stability is reached, typically declining by no more than  3\%. On CIFAR-10, the gap remains modest at larger channel counts — 1.5\% to 2.5\%. ViTs show the strongest sensitivity at a small width on all datasets. The largest differences appear in the unstable region and shrink as the width increases on MNIST and Fashion MNIST. Yet, a clear gap remains on CIFAR 10 even at the largest width in our sweep, about 4 to 6 percentage points.

On the most popular and widely used image classification tasks and across the three network families--MLP/DNNs, CNNs, and ViTs--a consistent picture emerges. The most reliable marker of stable learning is the consistency of accuracy after minimal learnable parameters have been reached. For example, 1000+ parameters in MLPs, irrespective of MLPs' depth, corroborate the universal approximation theorem, and \~4 channels in CNNs, \~6 hidden dimensions in ViT for MNIST. We observed that the order of minimal architecture increases proportionally to the complexity of the datasets. The practical boundary for safe compression for minimal architecture using pruning and quantization also follows a similar trend. We observe that larger architectures exhibit a more stable decline in performance with increasing pruning ratio than models with fewer parameters. This is attributed to the parameter redundancy apparent in networks. For 8-bit quantization, we observe that quantized models perform slightly worse than unquantized models; however, on more complex datasets like CIFAR-10, this gap widens. The results indicate that pruning and quantization for model compression should be applied only after convergence stability has been achieved. The convergence effectiveness is determined by how well architectural decisions match the task complexity: CNNs' architecture reaches stability with few channels because locality and weight sharing provide strong inductive bias, whereas the transformer requires a wider embedding to learn both local and global relations from data. The same order of operations holds across architectures. This analysis suggests that networks need to be trained to a stable convergence before pruning, and pruning will result in more graceful failure for a more stable minimal architecture than for smaller networks. Applying 8-bit quantization to a stable minimal architecture will result in minimal performance degradation. However, for more complex tasks, it is preferable to use a larger number of nodes/channels/hidden dimensions.

\section{Conclusions}\label{sec:conclusion}
We analyzed DNNs, CNNs, and ViTs in order to identify a minimal set of neural architectures capable of effectively solving the most widely known image classification tasks in the literature. Our analysis aimed to systematically establish a relation among the profiles of convergence dynamics, pruning, and quantization. Our initial investigation into various hidden-layer shapes revealed that DNN performance is invariant to network configuration when the number of parameters is sufficiently large. Consequently, we focused on networks with uniform layer sizes to isolate the influence of network configurations and extend the study to CNNs and ViT networks. Our experiments demonstrate that network performance exhibits {\it distinct learning regimes} ({\it unstable, stable, and over-fitting}) as the number of learnable parameters increases. Networks with lower parameter orders tend to be unstable and highly sensitive to optimization perturbations, while those in the ranges achieve stable and optimal performance. In the context of pruning, we found that pruning is a function of layers/channels/hidden dimensions. That is, deeper architectures can tolerate over $60\%$ sparsity with minimal loss in accuracy. 
%
%The quantization-aware training results indicate that quantization is a function of data complexity, and larger networks suffer from greater precision loss than shallower ones. 
%
% Giuseppe Nicosia: we should change the above sentence
%
Quantization sensitivity is a function of data complexity; while the gap between full and $8-bit$ precision decreases as network capacity moves into the stable region, a notable gap remains on the most complex datasets (CIFAR-10) even for larger, stable architectures.
However, quantizing ViT improved performance for some dimensions, opening a promising research direction to investigate the relationship between its components and quantization. Our integrated framework provides insights into the interplay between network structure and network compression techniques and lays the groundwork for developing minimal and resource-efficient neural models. %Future research will extend these findings to more complex deep-learning models and tasks, further advancing the quest for minimal yet high-performing networks.

% \section*{Acknowledgment}

\bibliographystyle{IEEEtran} % IEEE样式
\bibliography{refs} % 指定.bib文件名（不带扩展名）

%\vspace{12pt}
% \color{red}
% IEEE conference templates contain guidance text for composing and formatting conference papers. Please ensure that all template text is removed from your conference paper prior to submission to the conference. Failure to remove the template text from your paper may result in your paper not being published.

\end{document}